\pdfoutput=1

\documentclass[11pt]{article}

\usepackage{header}

\usepackage{emnlp2021}

\usepackage{times}
\usepackage{latexsym}

\usepackage[T1]{fontenc}

\usepackage[utf8]{inputenc}

\usepackage{microtype}

\usepackage{inconsolata}

\usepackage{microtype}
\usepackage{arydshln}
\usepackage{booktabs}
\usepackage{graphicx}
\usepackage{subcaption}
\usepackage{xspace}
\usepackage{amsmath}
\usepackage{cleveref}
\usepackage{parskip}
\usepackage{amsmath}
\usepackage{multirow}
\usepackage{threeparttable}
\usepackage{soul}
\usepackage{hyperref}

\newcommand{\projectAcro}{TOASTS\xspace}

\definecolor{blue}{RGB}{132, 160, 215}
\definecolor{green}{RGB}{189, 220, 171}

\DeclareRobustCommand{\bluehl}[1]{\sethlcolor{blue}{{\hl{~#1~}}}}
\DeclareRobustCommand{\greenhl}[1]{\sethlcolor{green}{{\hl{~#1~}}}}

\title{Analyzing Multi-Task Learning for Abstractive Text Summarization}

\author{Frederic Kirstein\textsuperscript{1,2,3}, Jan Philip Wahle\textsuperscript{1}, Terry Ruas\textsuperscript{1}, Bela Gipp\textsuperscript{1} \\
  \textsuperscript{1}Georg-August-Universität Göttingen, Germany \\
  \textsuperscript{2}Mercedes-Benz Group AG, Germany \\
\textsuperscript{3}\texttt{kirstein@gipplab.org} }
\begin{document}
\maketitle

\thispagestyle{firststyle}

\begin{abstract}
Despite the recent success of multi-task learning and pre-finetuning for natural language understanding, few works have studied the effects of task families on abstractive text summarization.
Task families are a form of task grouping during the pre-finetuning stage to learn common skills, such as reading comprehension.
To close this gap, we analyze the influence of multi-task learning strategies using task families for the English abstractive text summarization task.
We group tasks into one of three strategies, i.e., sequential, simultaneous, and continual multi-task learning, and evaluate trained models through two downstream tasks.
We find that certain combinations of task families (e.g., advanced reading comprehension and natural language inference) positively impact downstream performance. 
Further, we find that choice and combinations of task families influence downstream performance more than the training scheme, supporting the use of task families for abstractive text summarization.
Our code is publicly available \footnote{\url{https://github.com/FKIRSTE/GEM_emnlp2022-TOASTS}}.
\end{abstract}

\section{Introduction}
Self-supervised learning has been a significant success driver for generating high-quality abstractive summaries \cite{devlin-etal-2019-bert,https://doi.org/10.48550/arxiv.1907.11692, 10.1145/3416063,LewisLGG20,JMLR:v21:20-074,Radford2019}. 
Through self-supervision, language models implicitly learn intrinsic language features (e.g., syntax) from unlabeled data that they can use to solve downstream tasks \cite{NEURIPS2020_1457c0d6}. 
However, skills necessary to perform specific tasks often can be learned from an existing set of labeled data, requiring fewer training iterations \cite{rajpurkar-etal-2016-squad, see-etal-2017-get}.
For example, to perform text summarization, a helpful skill is the ability to answer questions about texts \cite{rajpurkar-etal-2016-squad}.

The multi-task learning paradigm and its variations aim to acquire multiple skills simultaneously to succeed on the downstream tasks, e.g., T5 \cite{JMLR:v21:20-074}, and are independent of a specific training stage \cite{aribandi2022ext}. 
While studies on the effects of multi-task learning on a large scale exist \cite{aghajanyan-etal-2021-muppet, Sun_Wang_Li_Feng_Tian_Wu_Wang_2020, aribandi2022ext} and are evaluated on broad natural language understanding benchmarks \cite{NEURIPS2019_4496bf24}, they are lacking insight on the influence on abstractive text summarization.
Furthermore, multi-task learning approaches are diverse in their methods (e.g., training scheme, mixing strategy, task families), hampering their comparison.

In this work, we investigate the role of multi-task learning on English abstractive text summarization. 
Therefore, we organize 18 pre-selected training tasks into six higher-level, modular task families.
Further, we compare three training schemes for the pre-finetuning stage and their respective mixing strategies through changes of multiple scores.

Our experiments show that families' choice significantly impacts text summarization, while different training schemes have little influence.
Moreover, pairing a text summarization task family with any other helps to stabilize the overall performance when transferring to unknown data.
In some cases, we also found that a text summarization task family can be substituted by other family pairs, e.g., advanced reading comprehension and classification. 

To summarize our contributions:
\begin{itemize}
    \item We study the influence of multi-task learning by training models on six task families for the English abstractive text summarization task.
    \item We evaluate the co-training of different task families using statistical (e.g., ROUGE) and semantic metrics (e.g., BERTScore) for 18 datasets.
    \item We compare the influence of three training schemes (i.e., sequential, simultaneous, continual multi-task learning) and two mixing strategies (i.e., proportional, equal).
\end{itemize}

\begin{table*}[htbp]
\centering
\small
\resizebox{\textwidth}{!}{
    \begin{tabular}{lllll}
    \toprule
\textbf{Task Family}                & \textbf{Task}         & \textbf{Dataset}         & \textbf{Source}             & \textbf{Characteristics}      \\ 
\toprule
Classification                      & Sentiment Classification & GoEmotion (\citeyear{demszky-etal-2020-goemotions})                 & Reddit                & multi-label CLS        \\
{[}CLS{]}                           & Sentiment Classification & IMDB (\citeyear{maas-EtAl:2011:ACL-HLT2011})                     & IMDB                  & binary CLS       \\
                                    & Topic Classification & AG News (\citeyear{https://doi.org/10.48550/arxiv.1509.01626})                  & ComeToMyHead          & multi-class CLS \\
\midrule
Commonsense                         & Fill-In-The-Blank     & Winogrande (\citeyear{10.1145/3474381})               & WSC dataset           & binary options \\
{[}CMNS{]}                          & Question Answering    & PhysicaliQA (\citeyear{https://doi.org/10.48550/arxiv.1911.11641})              & instructables.com     & binary options    \\
                                    & Question Answering    & SocialiQA (\citeyear{https://doi.org/10.48550/arxiv.1904.09728})                & crowdsourced          & ternary options \\
\midrule
Natural Language Inference          & Textual Entailment CLS & MNLI (\citeyear{N18-1101})                     & SNLI corpus           & multi-label CLS \\
{[}NLI{]}                           & Textual Entailment CLS & ANLI (\citeyear{https://doi.org/10.48550/arxiv.2010.12729})                     & \begin{tabular}[c]{@{}l@{}} human-and-model-\\in-the-loop dataset \end{tabular}         &  multi-label CLS \\
                                    & Textual Entailment CLS & QNLI (\citeyear{wang-etal-2018-glue})                      & Wikipedia             & binary classification \\
\midrule
Reading Comprehension               & Binary QA             & BoolQ (\citeyear{clark-etal-2019-boolq})                    & Google                & yes/no answer             \\
{[}RC{]}                            & Extractive QA         & SQuAD (\citeyear{rajpurkar-etal-2016-squad})                    & Wikipedia             & extractive answers          \\
                                    & Abstractive QA        & TweetQA (\citeyear{xiong-etal-2019-tweetqa})                   & Twitter               & abstractive answers allowed    \\
\midrule
Advanced RC                         & RC + Information Retrieval & HotpotQA (\citeyear{yang-etal-2018-hotpotqa})            & Wikipedia             & multi-hop question answering \\
{[}RC\textsuperscript{+}{]}         & RC + Open Domain QA    & Natural Questions (\citeyear{kwiatkowski-etal-2019-natural})         & Google, Wikipedia     & answer information seeking questions \\
                                    & RC + CMNS & ReCoRD (\citeyear{https://doi.org/10.48550/arxiv.1810.12885})                   & \begin{tabular}[c]{@{}l@{}} CNN/DailyMail \\ and Internet Archive \end{tabular}     & extractive Machine RC    \\
\midrule
Summarization                       & Extractive SUM        & XSum (\citeyear{xsum-emnlp})                     & BBC                   & one-sentence summary            \\
{[}SUM{]}                           & Abstractive SUM       & WikiLingua {[}eng{]} (\citeyear{ladhak-etal-2020-wikilingua})     & WikiHow               & one-sentence summary    \\
                                    & Abstractive SUM       & AESLC (\citeyear{zhang2019email})                    & E-Mail                & subject line generation     \\
\bottomrule
\end{tabular}
}
    \caption{Our selection of 18 representative datasets organized by their task family. For every dataset, we list the target task, the source, and the characteristics of the data. For a complete list of tasks, please see \Cref{ap:nlp_tasks}.}
    \label{tab:pre-finetuning_selection}
\end{table*}

\section{Related Work}
\textbf{Multi-task learning and pre-finetuning.} Transformers \cite{NIPS2017_3f5ee243} such as BERT \cite{devlin-etal-2019-bert} and GPT-3 \cite{NEURIPS2020_1457c0d6} are trained using a two-step approach, the \textit{pre-training} on large unlabeled corpora and the \textit{finetuning} on a smaller, more specific (and usually labeled) downstream corpus.
This bilateral approach allows language models to obtain general text representations once to perform many NLP downstream tasks with few gradient steps (e.g., document classification \cite{OstendorffRBG20,OstendorffRSR20}, plagiarism detection \cite{WahleRMG21a, WahleRFM22, wahle2022large}, media bias detection \cite{SpindePKR21,SpindeKRM22}). 
However, pre-training is typically highly computationally expensive and requires dedicated ample infrastructure; few researchers can reproduce the pre-training of large language models.
Therefore, recent works \cite{DBLP:journals/corr/abs-1811-01088, aghajanyan-etal-2021-muppet}) proposed additional training stages between pre-training and finetuning, i.e., \textit{pre-finetuning}\footnote{In this paper, we will use \textit{intermediate training} and \textit{pre-finetuning} interchangeably}.

ERNIE 2.0 \cite{Sun_Wang_Li_Feng_Tian_Wu_Wang_2020} proposes \textit{continual multi-task learning}, in which tasks are trained incrementally, thereby building a queue of introduced tasks that re-appear throughout the training process, to counter catastrophic forgetting \cite{MCCLOSKEY1989109,kirkpatrick2017overcoming}.
MUPPET \cite{aghajanyan-etal-2021-muppet} and ExT5 \cite{aribandi2022ext} follow a \textit{simultaneous} approach, drawing heterogeneous batches from multiple tasks and massively scale their training to >50 and >100 tasks respectively. 
MT-DNN \cite{liu-etal-2019-multi} organizes the prediction layer of a Transformer into four task families of common tasks of the GLUE benchmark \cite{wang-etal-2018-glue} and learns each task \textit{sequential} with their task order randomized.
This study compares continual multi-task learning, simultaneous training, and sequential training for abstractive text summarization.
 
\textbf{Task selection and relationship.}
\citet{vu-etal-2020-exploring} conduct an empirical investigation on 33 tasks across three broad groups (i.e., text classification, question answering, and sequence labeling) to explore their inter- and intra-group training for different group sizes.
Their experiments suggest that positive transfers between task groups are possible when the source dataset is small, and inter-group transfers are sensitive to group sizes.
ExT5 \cite{aribandi2022ext} analyzes the correlation of task family representatives and shows, that summarization tasks (i.e., CNN/Daily Mail \cite{see-etal-2017-get}, XSum \cite{xsum-emnlp}, WikiLingua\cite{ladhak-etal-2020-wikilingua}) generally reduce performance on most other task families and that CBQA tasks (i.e., Natural Questions \cite{kwiatkowski-etal-2019-natural}, Trivia QA \cite{JoshiTriviaQA2017}, Hotpot QA \cite{yang-etal-2018-hotpotqa}) are sensitive to multi-task learning.
For the task relationship and transfer analysis, \citet{aribandi2022ext} train on two families simultaneously and evaluate the first one.
We expand the study of \citet{aribandi2022ext} by adapting task families and respective representative tasks to be related to the text summarization task (\Cref{ssec:task_family_setup}), considering different family combinations, training approaches (\Cref{ssec:train_strategy}), and tracking their performance through additional metrics for different unseen datasets (\Cref{sec:experimental_setup}).

Multiple works leverage algorithms for the selection of training tasks, e.g., \citet{ruder-plank-2017-learning} use Bayesian Optimization to learn similarity measures (i.e., Jensen-Shannon divergence \cite{61115} and R\'{e}nyi divergence \cite{renyi1961measures}) and a Beta-Bernoulli multi-armed bandit with Thompson Sampling \cite{10.1561/2200000070, 10.1093/biomet/25.3-4.285} is used by AutoSem \cite{guo-etal-2019-autosem}.
Conversely, ExT5 \cite{aribandi2022ext} does not rely on automatic training task selection approaches as described by the preceding works and instead chooses an empirical approach to select tasks for higher-level task families.
We follow the approach of \citet{aribandi2022ext}'s task representative selection when choosing our tasks as the training task correlation analysis in ExT5 indicates which families could positively influence text summarization.

\section{Methodology}
\begin{figure*}
\small
    \centering
    \includegraphics[scale=.46]{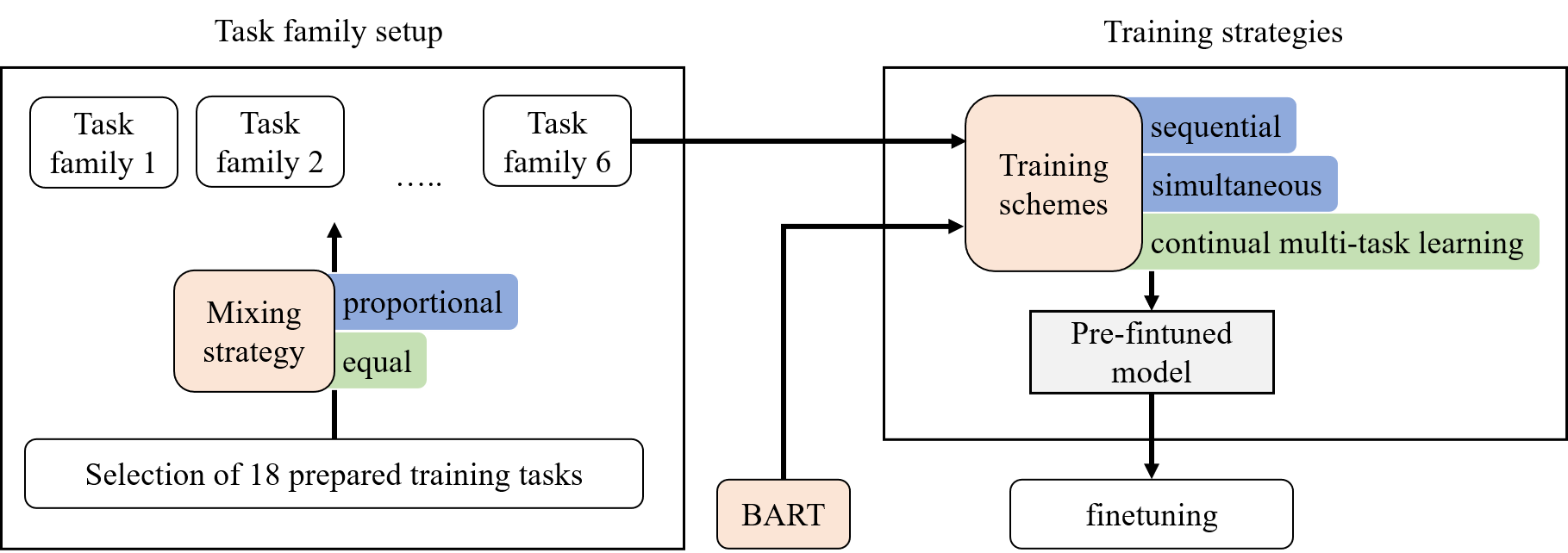}
    \caption{
    The central architecture of \projectAcro.
    The intermediate training phase commences the \textbf{task family setup} (left) by organizing the pre-selected training tasks into families of similar problems and applying two (proportional, equal) intra-family mixing strategies.
    The \textbf{training strategies} (right) continue by processing and organizing the generated task families into batches according to one of three training schemes (sequential, simultaneous, continual multi-task learning).
    After pre-finetuning BART, the resulting model is finetuned and evaluated on two abstractive text summarization datasets (Reddit TIFU, arXiv).
    The training/mixing scheme pairings are marked by the background colors \greenhl{green} and \bluehl{blue}.
    }
    \label{fig:overview_TOASTS}
\end{figure*}

We name our study \projectAcro, a \textbf{T}ask-\textbf{O}riented \textbf{A}nalysi\textbf{S} for \textbf{T}ext \textbf{S}ummarization to investigate the effects of different task family combinations on English abstractive text summarization via a multi-task learning architecture.
\projectAcro groups selected pre-training tasks into task families and explores the correlation of these families, their influence on two downstream tasks, and their aggregation through three training schemes.
Therefore, we use pre-finetuning, a second inexpensive pre-training stage between pre-training and fine-tuning, which was recently proposed by Muppet \cite{aghajanyan-etal-2021-muppet} and tested by ExT5 \cite{aribandi2022ext}.
Pre-finetuning has two main parts: the \textit{task family setup} and the \textit{training strategies}.
The task family setup groups different tasks and related datasets into broader families according to their primary objective.
The tasks of these families are then combined following a training strategy and evaluated into a final task. 
Figure \ref{fig:overview_TOASTS} illustrates the components of \projectAcro, which are detailed in the following sections.

\subsection{Task family setup}
\label{ssec:task_family_setup}
\noindent
\textbf{Selection.}
A myriad of NLP downstream tasks (e.g., word sense disambiguation and paraphrase detection) can be considered when choosing a multi-task architecture.
Without computational limits, one could explore all possible permutations of tasks and the influence of the respective tasks on downstream performance.
Unfortunately, as the number of tasks grows by more than their factorial number, joint training becomes computationally prohibitive \cite{aribandi2022ext}.
Therefore, we organize tasks into six high-level families \cite{aribandi2022ext, NEURIPS2020_1457c0d6} and perform combinations on their family levels: \textit{classification (CLS)}, \textit{commonsense reasoning (CMNS)}, and \textit{natural language inference (NLI)}, \textit{reading comprehension (RC)}, \textit{advanced reading comprehension \footnote{\citet{aribandi2022ext} refer to this family as Closed Book Question Answering (CBQA).} (RC\textsuperscript{+})}, \textit{summarization (SUM)}.
We compose each task family of three datasets that tackle different aspects of the problem, as shown in \Cref{tab:pre-finetuning_selection}.

The selected tasks in \projectAcro should not be seen as an exhaustive list of all NLP downstream tasks; instead, they should be considered an educated selection to measure task family influence on text summarization.
An extended list of planned tasks for future analyses can be found in \Cref{tab:full_list} in \Cref{ap:nlp_tasks}. 

\noindent
\textbf{Task mixing.}
After pre-selecting representative tasks for each family, we control the percentage of data ingested from each task using a task mixing strategy.
We consider two methods for processing all combinations of task families: \textit{proportional mixing} \cite{10.1609/aaai.v33i01.33016949,aribandi2022ext} and \textit{equal mixing} \cite{JMLR:v21:20-074}.
Equal mixing picks training samples from each task with equal probability, while proportional mixing sets the probability to the proportion of each task's size.
The use of proportional mixing as a default strategy is the recommended approach for various multi-task learning strategies \cite{10.1609/aaai.v33i01.33016949}.
However, continual multi-task learning (\Cref{ssec:train_strategy}) requires an equal mixing strategy even though related studies have shown it to be sub-optimal \cite{JMLR:v21:20-074}.
While we sample either proportional or equal within task families, we draw equal between task families to balance the influence of potentially different task families.
We leave to future work the investigation of the effects of different amounts of tasks and samples per family.

\subsection{Training strategies} \label{ssec:train_strategy}
\begin{figure*}[htbp]
    \centering
    \begin{subfigure}{.25\textwidth}
        \centering
        \includegraphics[scale=.5]{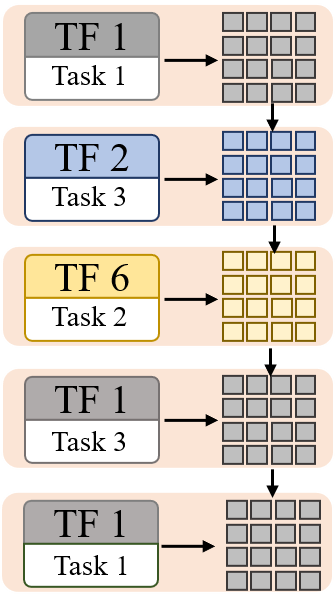}
        \caption{Sequential learning.}
    \end{subfigure}%
    \begin{subfigure}{.25\textwidth}
        \centering
        \includegraphics[scale=.5]{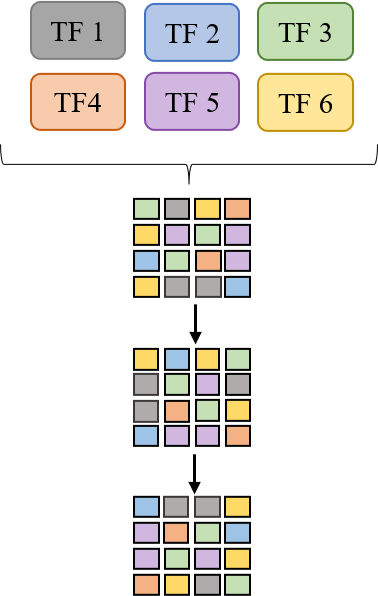}
        \caption{Simultaneous learning.}
    \end{subfigure}%
    \begin{subfigure}{.5\textwidth}
        \centering
        \includegraphics[scale=.5]{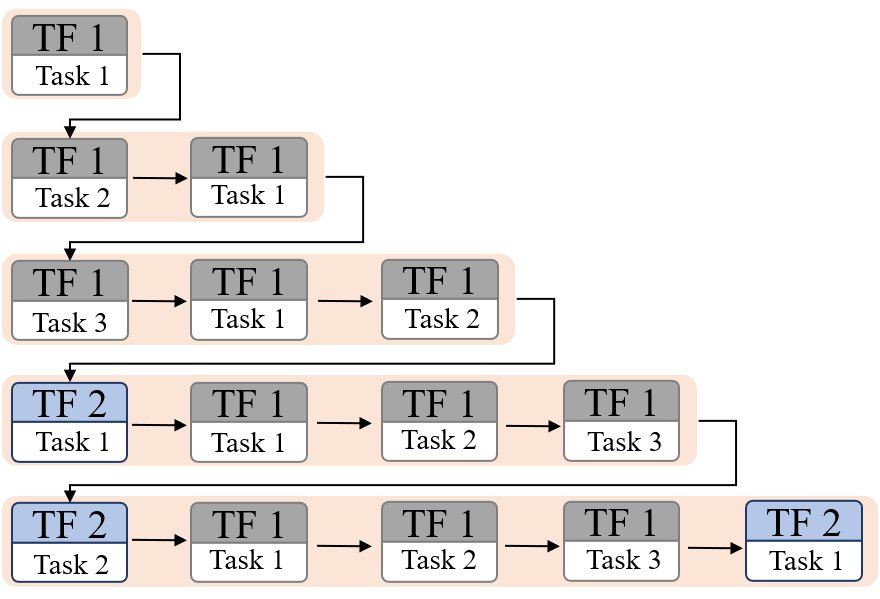}
        \caption{Continual multi-task learning.}
    \end{subfigure}%
    \caption{\projectAcro's three training strategies. (a) Sequential learning (seq) draws a batch with samples from one task of a task family at a time for every training stage. The order of tasks is randomized. (b) Simultaneous learning (sim) samples from all available tasks at the same time. (c) Continual multi-task learning (cMTL) introduces a new task in each training stage, which is added to the end of the training queue.}
     \label{fig:training_schemes}
\end{figure*}

\noindent
\textbf{Training Schemes.}
Multi-task learning during a pre-finetuning stage allows us to start from a pre-trained checkpoint, decreasing the final task's overall cost.
We explore three multi-task learning training schemes for the pre-finetuning as \Cref{fig:training_schemes} shows: \textit{sequential learning} (seq) \cite{MCCLOSKEY1989109,biesialska-etal-2020-continual}, \textit{simultaneous learning} (sim) \cite{10.1023/A:1007379606734, aghajanyan-etal-2021-muppet}, and \textit{continual multi-task learning} (cMTL) \cite{Sun_Wang_Li_Feng_Tian_Wu_Wang_2020}. 
In the \textit{sequential} approach, training batches are composed of a single dataset, i.e., homogeneous batches, and their processing order is sequentially randomized \cite{liu-etal-2019-multi}.
This approach achieves a concentrated task learning on the batch level while keeping the overall variety, therefore learning a task more thoroughly before moving to the next.
For the \textit{simultaneous} strategy, we combine all tasks into a single pool and draw randomly from it \cite{aghajanyan-etal-2021-muppet, aribandi2022ext}.
This prominent approach introduces task variety on the batch level by constantly challenging the model with different approaches, forcing it to identify intrinsic commonality between the task families quickly.
For \textit{continual multi-task learning}, we adjust the concept of ERNIE 2.0 \cite{Sun_Wang_Li_Feng_Tian_Wu_Wang_2020} to adapt it to our task family configuration.
As our tasks corpus is less extensive than the training dataset used in ERNIE 2.0, we have to rejig the number of stages and training steps in \projectAcro.
Therefore, when including new tasks and task families, we change their total number of steps to 9k, and 27k, respectively, as \Cref{tab:train_strategy_details} shows.
One difference from ERNIE 2.0 is that once a new task is introduced to the pipeline and trained for the first time at timestep $t$, we move it to the end of the queue of previously trained tasks as the last one to be executed in $t+1$.
Using the order in \cite{Sun_Wang_Li_Feng_Tian_Wu_Wang_2020} as an alternative way of including and carrying new tasks, yields worse results (\Cref{tab:cl_asc_VS_cl_desc}).
Through the pre-determined task order of this approach, we can control which task families follow each other and how fundamental a task is by introducing it earlier than others. 

\begin{table}[htbp]
  \centering
  \small
    \begin{tabular}{llllllll}
    \toprule
    \textbf{Task} & \textbf{S1} & \textbf{S2} & \textbf{S3} & \textbf{S4} & \textbf{S5} & … &\textbf{S18}\\
    \cmidrule(lr){1-1}\cmidrule(lr){2-8}
    TF1.1 & 500   & 500   & 500   & 500   & 500   & … & 500\\
    TF1.2  &    -   & 1k  & 500   & 500   & 500   & … & 500\\
    TF1.3 &   -    &   -    & 1.5k  & 500   & 500   & … & 500\\
    TF2.1  &   -    &   -    &   -    & 2k  & 500   & … & 500\\
    TF2.2 &   -    &    -   &    -   &   -    & 2.5k  & … & 500\\
    …     &  -    &    -   &    -   &   - &   -   & … & 500\\
    TF6.3 &  -    &    -   &    -   &   - &   -   & - & 9k\\
    \bottomrule
    \end{tabular}%
    \label{tab:addlabel}%
    \caption{
    The number of batches during cMTL training depends on the training stage and the number of introduced tasks. S1 to S16 denote the stages when a new task TF1.1 to TF6.3 is introduced. TF1.1 indicates the first task of task family 1, TF1.2 the second task of task family 1 etc.
    }
    \label{tab:train_strategy_details}
\end{table}%

\section{Experimental setup} \label{sec:experimental_setup}
\noindent
\textbf{Model.}
For all experiments, we use BART-{Large} \cite{LewisLGG20} to probe combinations of task families, mixing, and training strategies in \projectAcro.
BART is a two-stage denoising autoencoder that corrupts its input text and reconstructs it through a sequence-to-sequence model.
We chose BART because of its ability to perform a wide range of downstream tasks, such as paraphrase detection \cite{WahleRFM22}, fake news identification \cite{WahleARM22}, and text summarization \cite{LewisLGG20}.
Additionally, in our preliminary experiments, BART also performed better than other candidate models such as PEGASUS \cite{ZhangZSL20} and T5 \cite{JMLR:v21:20-074} (comparison in \Cref{tab:model_comparison_reddit,tab:model_comparison_arxiv} in \cref{ap:more models}). 

\noindent
\textbf{Tokenization.} 
We tokenize text using the BART-{Large} tokenizer and augment all texts to include task-specific prompts such as \texttt{'question:'} or \texttt{'context:'}.
Further, we structure the samples to follow a uniform text-to-text style which allows the model to handle multi-task learning across different task families without needing task-specific losses, loss scaling or explicit gradient accumulation on heterogeneous batches \cite{liu-etal-2019-multi, aghajanyan-etal-2021-muppet}.

\noindent
\textbf{Hyperparameters.}
We run our experiments on 8 NVIDIA A100s with a total of 320GB GPU memory.
The models are trained with a total batch size of 8 for three epochs and up to 60k global steps for six task families during pre-finetuning (finetuning: 16k for Reddit TIFU, 70k for arXiv) with half-precision (fp16).
The pre-finetuning takes between 17min (single task family) and 11h (all task families).
The finetuning takes 2.2h for Reddit TIFU and 19.85h for arXiv.
During pre-finetuning, we set the input sequence to 512 tokens and the target sequence to 128 as a compromise for training time and context.
During finetuning, the sequence lengths are increased to 1024 and 512 for input and target, respectively, to capture the full context of both evaluation datasets.
For other hyperparameters we refer the reader to \Cref{tab:hyperparameters} in \Cref{ap:hyperparameters}.

\begin{table*}[htbp]
\small
  \centering
    \begin{tabular}{l ccc ccc}
    \toprule
    \multicolumn{1}{c}{\multirow{2}[4]{*}{\textbf{Task Families}}} & \multicolumn{3}{c}{\textbf{Reddit TIFU}} & \multicolumn{3}{c}{\textbf{arXiv}} \\
\cmidrule(lr){2-4}  \cmidrule(lr){5-7}       
& seq   & sim   & cMTL  & seq   & sim   & cMTL \\
\midrule
    CLS   & 0.226 & 0.233 & 0.060 & 0.154 & 0.287 & \textbf{0.286} \\
    CMNS  & 0.226 & 0.078 & 0.078 & 0.286 & 0.197 & 0.163 \\
    NLI   & 0.030 & 0.082 & 0.082 & 0.168 & 0.111 & 0.182 \\
    RC    & 0.230 & \underline{\textbf{0.235}} & 0.230 & 0.282 & 0.284 & 0.282 \\
    RC\textsuperscript{+}  & 0.224 & 0.082 & 0.078 & 0.282 & \textbf{\underline{0.289}} & 0.203 \\
    SUM   & \textbf{0.231} & \underline{\textbf{0.235}} & \textbf{0.231} & \textbf{0.288} & 0.282 & \textbf{0.286} \\
    ALL   & 0.222 & 0.228 & 0.037 & 0.281 & 0.279 & 0.008 \\
    \midrule
    BART (baseline)  & 0.087$^\dagger$ & 0.087$^\dagger$ & 0.087$^\dagger$ & 0.281$^\dagger$ & 0.281$^\dagger$ & 0.281$^\dagger$ \\
    \bottomrule
    \end{tabular}%
    \caption{Results (METEOR) for single task families and the combination of all task families for the Reddit TIFU and arXiv datasets. Values in \textbf{bold} represent the highest results for a training scheme. \underline{Underlined} values are the highest results for that dataset independent from training. $^\dagger$Repeated result for baseline without training scheme.}
  \label{tab:RQ1_meteor_all}%
\end{table*}%

\noindent
\textbf{Evaluation.}
To understand each task family's influence, mixing, and training strategies, we evaluate the text summarization task using two datasets: Reddit TIFU \cite{kim-etal-2019-abstractive} and arXiv \cite{cohan-etal-2018-discourse}.
Reddit TIFU is composed of 120K posts from online conversations, with the task of creating a tldr\footnote{too long; didn't read} summary from the post.
The arXiv dataset consists of 250K scientific articles with the task of deriving the abstract from the full text.
These datasets are commonly referred to as challenging abstractive summarization tasks \cite{ZhangZSL20,HeKMR2020}.
In combination, they provide a balanced landscape as Reddit TIFU contains shorter examples with an average of 432 words per post and 23 per summary, relying on simpler linguistic, and arXiv longer examples with 4938 words per document and 220 per summary constructed from elaborated text.

During our experiments, we consider a combination of \textit{count-based} and \textit{semantic} metrics to assess the quality of produced summaries.
We use BLEU \cite{papineni-etal-2002-bleu}, ROUGE (1, 2, L) \cite{lin-2004-rouge}, and METEOR \cite{banerjee-lavie-2005-meteor}, which favor precision, recall, and harmonic mean, respectively.
Even though these traditional metrics can work well for similarly worded summaries, they are limited when wording changes, but the semantic meaning remains the same \cite{bhandari-etal-2020-evaluating, huang2021factual}.
To assess semantic similarity better, we also include BERTScore \cite{DBLP:journals/corr/abs-1904-09675}, a similarity measure that maximizes the cosine similarity between candidate and reference contextualized token embeddings via BERT \cite{devlin-etal-2019-bert} in a greedy manner.

\subsection{Experimental results and discussion}
We structure our experiments into four research questions, which tackle the relevance of task families and dataset compatibility (RQ1), the effects of co-training text summarization task families with other families (RQ2), the co-training of task families excluding text summarization (RQ3), and the co-training of text summarization and two different task families (RQ4).

We pre-finetune our baseline model (BART-Large) for each experiment on specific task families (e.g., CLS, CMNS) and evaluate the resulting models into the Reddit TIFU and arXiv datasets.
\Crefrange{tab:RQ1_meteor_all}{tab:RQ4_meteor_reddit} show the different task mixing and training strategies.
Sequential (seq) and simultaneous (sim) training strategies use proportional mixing, while continual multi-task learning (cMTL) uses equal mixing. 
Because of space constraints, we report our results only for the METEOR metric, which proved to be the most sensitive to our experiments. 
We include a complete list of results for BertScore, BLEU, METEOR, and ROUGE (1, 2, L) in \Cref{ap:complete_results_reddit,ap:complete_results_arxiv}.

\textit{\textbf{RQ1}: Does increasing the number of pre-finetuning datasets increase downstream task performance for text summarization?}\\[1.5pt]
A. To identify if the text summarization downstream task benefits from unconstrained usage of multiple task families, we compare how each task family performs against the combination of all.

As \Cref{tab:RQ1_meteor_all} shows, the SUM task family consistently outperforms the combination of all families for both datasets (followed by RC), except for the sim training scheme on arXiv.
The increase in performance through pre-training SUM is somehow expected, as it is the most related task family to the actual problem, i.e., abstractive text summarization.
Conversely, NLI performs the worst when compared to any other task family.
Pre-finetuning generally positively affects BART compared to its baseline, except for a few cases (e.g., cMTL-RC\textsuperscript{+}, NLI).
Overall, the sim training strategy greatly influenced downstream task performance.

Our results suggest that combining all task families is suboptimal for text summarization, which challenges recent observations for other NLP tasks \cite{aghajanyan-etal-2021-muppet, aribandi2022ext}. Also, increasing the number of task families requires high compute budgets.
As we train each task family individually or all simultaneously, it is unclear how much influence a summarization task family (e.g., SUM) has on the others.

\begin{table*}[htbp]
\small
  \centering
    \begin{tabular}{l ccc ccc}
    \toprule
    \multicolumn{1}{c}{\multirow{2}[4]{*}{\textbf{Task Families}}} & \multicolumn{3}{c}{\textbf{Reddit TIFU}} & \multicolumn{3}{c}{\textbf{arXiv}} \\
\cmidrule (lr){2-4} \cmidrule (lr){5-7}          
    & seq   & sim   & cMTL  & seq   & sim   & cMTL \\
\midrule  
    SUM+CLS & 0.230 & \textbf{0.233} & 0.077 & \textbf{0.285} & 0.285 & 0.283 \\
    SUM+CMNS & 0.232 & 0.231 & \textbf{\underline{0.234}} & 0.153 & 0.286 & \textbf{\underline{0.288}} \\
    SUM+NLI & 0.223 & \textbf{0.233} & 0.223 & 0.282 & \textbf{0.287} & 0.282 \\
    SUM+RC & \textbf{0.233} & 0.229 & \textbf{\underline{0.234}} & \textbf{0.285} & 0.280 & 0.283 \\
    SUM+RC\textsuperscript{+} & 0.230 & 0.225 & \textbf{\underline{0.234}} & 0.284 & 0.281 & 0.284 \\
    \midrule
    BART (baseline) & 0.087$^\dagger$ & 0.087$^\dagger$ & 0.087$^\dagger$ & 0.281$^\dagger$ & 0.281$^\dagger$ & 0.281$^\dagger$ \\
    \bottomrule
    \end{tabular}%
    \caption{Results (METEOR) for the combination of SUM and different task families for the Reddit TIFU and arXiv datasets. Values in \textbf{bold} represent the highest results for a training scheme. \underline{Underlined} values are the highest results for that dataset independent of training. $^\dagger$Repeated result for baseline without training scheme.}
  \label{tab:RQ2_meteor_all}%
\end{table*}%

\textit{\textbf{RQ2}: How much does the text summarization task affect other task families?}\\[1.5pt]
A. As SUM is closely related to the text summarization task, and it yields the best results in RQ1, we explore how its combination with another task family affects the resulting model.
\Cref{tab:RQ2_meteor_all} shows the results of combining SUM with other task families.
Aside from a few cases (e.g., arXiv sim for SUM+RC\textsuperscript{+}), pairing with the SUM family improves over almost every single run in \Cref{tab:RQ1_meteor_all} and the combination of all task families.

While some task families' combinations obtain small benefits (seq-SUM+RC), others are greatly affected (e.g., cMTL-SUM+CMNS) for both datasets.
The BART baseline performs better than the pre-finetuning in only two cases, i.e., SUM+CLS for Reddit TIFU (cMTL) and SUM+CMNS for arXiv (seq).
We observe fewer outliers with low scores when pairing SUM with other task families than in RQ1.
Individual training improved the performance on arXiv the most (seq and sim), while for Reddit TIFU, the combination of task families was more effective (seq and cMTL).

Low scores are also less frequent when combining task families with one exception, i.e., cMTL-SUM+CLS for Reddit TIFU.
The lowest scores in RQ1 (e.g., NLI, CMNS) and RQ2 (CLS) might be related to the fact that these tasks are not contributing to the learned weights of the downstream task.
As Reddit TIFU uses mostly informal language and its input sequence and summaries are short, this might justify these low scores.

The improvements in \Cref{tab:RQ2_meteor_all} over the BART baseline are likely to be related to the SUM family rather than a mixing strategy or training scheme.
The results of individually training the SUM family (RQ1) are equal or marginally higher when combined with other task families (e.g., 0.233 for SUM+RC vs. 0.231 SUM).
As the SUM family seems to substantially impact co-training multiple tasks, we are interested in evaluating the influence of families other than SUM.

\begin{table*}[htbp]
\small
  \centering
    \begin{tabular}{l ccc ccc}
    \toprule
    \multicolumn{1}{c}{\multirow{2}[4]{*}{\textbf{Task Families}}} & \multicolumn{3}{c}{\textbf{Reddit TIFU}} & \multicolumn{3}{c}{\textbf{arXiv}} \\
\cmidrule (lr){2-4} \cmidrule (lr){5-7}          
    & seq   & sim   & cMTL  & seq   & sim   & cMTL \\
\midrule
    CLS+CMNS & 0.078 & 0.078 & 0.060 & 0.078 & 0.050 & 0.162 \\
    CLS+NLI & 0.077 & 0.077 & 0.046 & 0.050 & 0.003 & 0.276 \\
    CLS+RC & 0.231 & 0.231 & 0.230 & \textbf{0.287} & 0.283 & 0.181 \\
    CLS+RC\textsuperscript{+} & 0.229 & 0.229 & 0.082 & 0.284 & 0.288 & \textbf{0.287} \\
    CMNS+NLI & 0.231 & 0.231 & 0.081 & 0.137 & 0.212 & 0.118 \\
    CMNS+RC & 0.227 & 0.227 & 0.077 & 0.283 & 0.284 & 0.179 \\
    CMNS+RC\textsuperscript{+} & 0.232 & 0.232 & \textbf{0.232} & 0.279 & 0.280 & 0.082 \\
    NLI+RC & 0.231 & 0.231 & 0.231 & 0.285 & 0.285 & 0.284 \\
    NLI+RC\textsuperscript{+} & \textbf{0.233} & \underline{\textbf{0.234}} & 0.227 & 0.286 & \textbf{\underline{0.290}} & 0.282 \\
    RC+RC\textsuperscript{+} & 0.228 & 0.228 & 0.228 & \textbf{0.287} & 0.281 & 0.285 \\
    \midrule
    BART (baseline)  & 0.087$^\dagger$ & 0.087$^\dagger$ & 0.087$^\dagger$ & 0.281$^\dagger$ & 0.281$^\dagger$ & 0.281$^\dagger$ \\
    \bottomrule
    \end{tabular}%
  \caption{Results (METEOR) for the combination of all pairs of task families (except for SUM) for the Reddit TIFU and arXiv datasets. Values in \textbf{bold} represent the highest results for a training scheme. \underline{Underlined} values are the highest results for that dataset independent of training. $^\dagger$Repeated result for baseline without training scheme.}
  \label{tab:RQ3_meteor_all}%
\end{table*}%
\textit{\textbf{RQ3}: How do non text summarization task families influence each other?}\\[1.5pt]
A. We remove the SUM family and co-train all possible pairs of task families.
\Cref{tab:RQ3_meteor_all} shows that the co-training of non text summarization task families (e.g., NLI+RC\textsuperscript{+}) can achieve equal or better results in comparison to single SUM training (\Cref{tab:RQ1_meteor_all}) or its combination with other task families (\Cref{tab:RQ2_meteor_all}) for both Reddit TIFU and arXiv.
Other combinations such as CLS+RC and RC+RC\textsuperscript{+} also achieve strong results.

Conversely, the combination of task families with good results individually seems to have a harmful influence on each other when paired.
While  CLS and CMNS have good results individually (0.226 and 0.226 for the seq strategy on Reddit  TIFU), their pairing (e.g., CLS+CMNS) is strongly negative (e.g., 0.078 for the seq strategy on Reddit TIFU).
As in \Cref{tab:RQ1_meteor_all}, different training schemes seem to be a less dominant factor than task family choice during pre-finetuning.
Therefore, a proper task family combination should precede architectural training options. 

Our results suggest that non text summarization task families can be used to substitute for the SUM family.
Specifically, all best-performing results include RC or RC\textsuperscript{+} in their configuration.
A possible explanation for the stark influence of RC/RC\textsuperscript{+} is that their problem of understanding texts is closely related to summarizing texts.
A link between reading comprehension and text summarization is also observed by psychologists in various studies (e.g., \citet{doi:10.1080/15434300701333129,Kintsch_1978,doi:10.1177/0265532208094275}).

\begin{table}[!htbp]
\small
  \centering
    \begin{tabular}{lccc}
    \toprule
    \multicolumn{1}{c}{\multirow{2}[3]{*}{\textbf{Task Families}}} & \multicolumn{3}{c}{\textbf{Reddit TIFU}} \\
\cmidrule(lr){2-4}          & seq   & sim   & cMTL \\
\midrule   
    {SUM+CLS+CMNS} & 0.228 & 0.227 & 0.077 \\
    {SUM+CLS+NLI} & 0.231 & 0.231 & 0.082 \\
    {SUM+CLS+RC} & \textbf{0.235} & 0.228 & 0.229 \\
    {SUM+CLS+RC\textsuperscript{+}} & \textbf{0.235} & 0.233 & 0.082 \\
    {SUM+CMNS+NLI} & 0.230 & \textbf{\underline{0.236}} & 0.229 \\
    {SUM+CMNS+RC} & 0.234 & 0.232 & \textbf{0.230} \\
    {SUM+CMNS+RC\textsuperscript{+}} & 0.232 & 0.231 & 0.228 \\
    {SUM+NLI+RC} & 0.229 & 0.231 & 0.228 \\
    {SUM+NLI+RC\textsuperscript{+}} & 0.234 & 0.229 & 0.229 \\
    {SUM+RC+RC\textsuperscript{+}} & 0.227 & 0.234 & 0.228 \\
    \midrule
    BART (baseline)  & 0.087$^\dagger$ & 0.087$^\dagger$ & 0.087$^\dagger$ \\
    \bottomrule
    \end{tabular}%
  \caption{Results (METEOR) for the combination of all pairs of task families and SUM for Reddit TIFU. Values in \textbf{bold} represent the highest results for a training scheme. \underline{Underline} values are the highest results for that dataset independent of training. $^\dagger$Repeated result for baseline without training scheme.}
  \label{tab:RQ4_meteor_reddit}%
\end{table}%

\textit{\textbf{RQ4}: How are non text summarization task family pairs affected by SUM?}\\[1.5pt]
A. Considering the positive effect of SUM in other families (RQ2), we investigate its influence in task family pairs (RQ3) as \Cref{tab:RQ4_meteor_reddit} shows.
For this research question, we only consider Reddit TIFU as it provides a more challenging scenario (i.e., informal, short texts) and limits our computational budget (family co-training is increasingly expensive when the number of task families grows).

Including SUM mitigates the adverse effects of combining CLS+CMNS (e.g., 0.228 vs. 0.078 for the seq training scheme) and CLS+NLI (e.g., 0.231 vs. 0.077 for the seq training scheme), except for the cMTL training scheme.
However, the scores for CLS+RC\textsuperscript{+} are almost unchanged.
The seq and sim training schemes still perform best (e.g., CMNS+NLI) but for different task family combinations compared to the previous research questions' results (e.g., NLI+RC\textsuperscript{+} in RQ3).
For the best performing task families pairs in RQ3, only CLS+RC and CLS+RC\textsuperscript{+} are still the top results when including SUM.
As in \Cref{tab:RQ2_meteor_all}, the SUM family seems to provide stability to the results, as we see fewer fluctuations than in \Cref{tab:RQ3_meteor_all}.
We assume the stability provided by SUM would also be present in the inclusion of more task families.
Further, we observe the positive influence of RC and RC\textsuperscript{+} when pairing three task families excluding SUM (\Crefrange{tab:RQ4_seq_wo_red}{tab:RQ4_cMTL_wo_red}).

\section{Conclusion \& Future Work}
In this work, we studied the influence of multi-task learning combinations of task families during the pre-finetuning stage for English abstractive text summarization. We trained three different training strategies, six task families composed of 18 tasks, and evaluated two downstream tasks.

Our experiments show that non text summarization task families, e.g., advanced reading comprehension, can be used as a substitute for the summarization task (RQ2) or the combination of all task families (RQ1).
However, including the summarization task family in the training process positively impacts the downstream performance compared to non text summarization family combinations.
Further, our analysis shows that training strategies have little influence on the overall performance compared to the task family selection.

We see this analysis as the first step to understanding training strategies and task families for text summarization.
In the future, we want to investigate more tasks (both in number and diversity) per task family, training schemes, and mixing strategies. 
We also plan to include psychological studies comparing the similarities of textual understanding tasks as a starting point for task family pre-selection.

\section*{Limitations}
With the organization of tasks and datasets into task families, this study highly depends on these representative tasks' domain and expressiveness.
As \citet{aribandi2022ext} faced similar problems, we followed their guidance to select representatives to consist of a diverse set of datasets to train and evaluate on and to partition task families as mutually exclusive as possible while being related to abstractive text summarization.
However, none of the datasets are perfectly isolated and can only be used as a proxy for a larger task family.

\section*{Ethical Considerations}
This study depends on existing resources and generative models; thus, it is not free of biases and possible ethical considerations.
One problem is the generation of text summaries that contain nonfactual information, meaning distortion, social biases such as political stances, or abusive language \cite{gooding-2022-ethical}.
To mitigate these problems we plan to condition the generation of trained models for unsafe content or other harmful text to return an empty string.

Furthermore, \projectAcro is licensed to the public under a copyright policy that allows unlimited reproduction, distribution, and hosting on any website or medium. 
Hence, anyone can exploit its limitations and inherited biases to propagate and amplify unintentional societal problems.

\bibliography{anthology.bib,custom}
\bibliographystyle{acl_natbib}

\appendix
\newpage
\section{Tasks and Families} \label{ap:nlp_tasks}
\Cref{tab:full_list} shows an extended version of pre-finetuning tasks in \Cref{tab:pre-finetuning_selection} to-be-considered in future work
\begin{table*}[htbp]
  \centering
  \small
    \begin{tabular}{llll}
    \toprule
    \textbf{TF} & \textbf{Task}  & \textbf{Dataset} & \textbf{Citation} \\
    \cmidrule(lr){1-4}
    CLS   & Topic Classification & AG News & \cite{https://doi.org/10.48550/arxiv.1509.01626}\\
          & Text Classification & Civil Comments & \cite{DBLP:journals/corr/abs-1903-04561}\\
          & Text Classification & FEVER & \cite{thorne-etal-2018-fever}\\
          & Emotion Classification & GoEmotions & \cite{demszky-etal-2020-goemotions}\\
          & Sentiment Classification & IMDB & \cite{maas-EtAl:2011:ACL-HLT2011}\\
          & Sentiment Classification & Rotten Tomatoes & \cite{Pang+Lee:05a}\\
          & Text Classification & Trec & \cite{li-roth-2002-learning,hovy-etal-2001-toward}\\
          & \multicolumn{1}{p{15.43em}}{classification} & Word-in-Context & \cite{https://doi.org/10.48550/arxiv.1808.09121}\\
          & Sentiment Classification & Yelp Polarity & \cite{https://doi.org/10.48550/arxiv.1509.01626}\\
          & Linguistic Acceptability & CoLA & \cite{warstadt2018neural}\\
          & Sentiment Classification & SST-2 & \cite{socher-etal-2013-recursive}\\
    \cmidrule(lr){1-4}
    CMNS  & Open Domain QA & AI2 Reasoning \\&&(Challenge ARC)  & \cite{Yadav_2019}\\
          & Concepts to Text Generation & CommonGen (CG) & \cite{https://doi.org/10.48550/arxiv.1911.03705}\\
          & Sequential Question Answering & CQA & \cite{1801.10314}\\
          & Commonsense Inference & HellaSWAG & \cite{zellers-etal-2019-hellaswag}\\
          & Question Answering & PhysicaliQA & \cite{https://doi.org/10.48550/arxiv.1911.11641}\\
          & Question Answering & SocialiQA & \cite{https://doi.org/10.48550/arxiv.1904.09728}\\
          & Text Classification & SWAG & \cite{zellers-etal-2018-swag}\\
          & Fill-In-A-Blank & WinoGrande & \cite{10.1145/3474381}\\
          & Question Answering & Winograd Scheme \\&&(Challenge) & \cite{https://doi.org/10.48550/arxiv.2004.13831}\\
          & Open-Domain-QA & CommonSense QA & \cite{https://doi.org/10.48550/arxiv.1811.00937}\\
    \cmidrule(lr){1-4}
    NLI   & Textual Entailment Classification & ANLI (Adverserial NLI) & \cite{https://doi.org/10.48550/arxiv.2010.12729}\\
          & Natural Language Inference & HANS & \cite{https://doi.org/10.48550/arxiv.1902.01007}\\
          & Textual Entailment Classification & MNLI & \cite{N18-1101}\\
          & Textual Entailment Classification & QNLI & \cite{wang-etal-2018-glue}\\
          & Textual Entailment Classification & RTE & \cite{https://doi.org/10.48550/arxiv.2010.03061}\\
          & Textual Entailment Classification & SciTail & \cite{10.5555/3504035.3504671}\\
          & Natural Language Inference & SNLI & \cite{https://doi.org/10.48550/arxiv.1909.02209}\\
          & Natural Language Inference & WNLI & \cite{wang-etal-2018-glue}\\
    \cmidrule(lr){1-4}
    RC    & Binary QA & BoolQ & \cite{clark-etal-2019-boolq}\\
          & Multiple Choice QA & Cosmos QA & \cite{https://doi.org/10.48550/arxiv.1909.00277}\\
          & Multi-Sentence QA & Eraser Multi RC & \cite{eraser2019,MultiRC2018} \\
          & Extractive QA & SQUAD & \cite{rajpurkar-etal-2016-squad}\\
          & Extractive QA & TriviaQA & \cite{JoshiTriviaQA2017}\\
          & Abstractive QA & TweetQA & \cite{xiong-etal-2019-tweetqa}\\
          & Multiple Choice QA & RACE & \cite{lai-etal-2017-race}\\
    \cmidrule(lr){1-4}
    RC\textsuperscript{+}   & Text2Text Generation & E2E & \cite{dusek.etal2020:csl}\\
          & RC + Question Answering & MSMarco & \cite{https://doi.org/10.48550/arxiv.1611.09268}\\
          & RC + Open Domain QA & Natural Questions & \cite{kwiatkowski-etal-2019-natural}\\
          & RC + Commonsense Reasoning   & RECORD & \cite{https://doi.org/10.48550/arxiv.1810.12885}\\
          & RC + Information Retrieval & HotpotQA & \cite{yang-etal-2018-hotpotqa}\\
          & RC + Extractive QA & DROP & \cite{Dua2019DROP}\\
    \cmidrule(lr){1-4}
    SUM   & Abstractive Summarization & Aeslc & \cite{zhang2019email} \\
          & Extractive Summarization & Billsum & \cite{Eidelman_2019}\\
          & Abstractive Summarization & CNN & \cite{see-etal-2017-get, DBLP:conf/nips/HermannKGEKSB15}\\
          & Headline Generation & Gigaword & \cite{https://doi.org/10.48550/arxiv.1509.00685}\\
          & Abstractive Summarization & Multinews & \cite{alex2019multinews}\\
          & Abstractive Summarization & WikiLingua {[}eng{]} & \cite{ladhak-etal-2020-wikilingua} \\
          & Extractive Summarization & XSUM & \cite{xsum-emnlp}\\

    \bottomrule
    \end{tabular}%
    \caption{An extended list of \Cref{tab:pre-finetuning_selection}. This list can be used to extend \projectAcro to more tasks and datasets in future work. TF stands for Task Family.}
    \label{tab:full_list}
\end{table*}

\section{Additional Models} \label{ap:more models}
\Crefrange{tab:cl_asc_VS_cl_desc}{tab:model_comparison_arxiv} shows the results for different models and loop orders. BART performed best compared to models from related work, which is why we chose the model throughout our experiments.
\begin{table*}[ht]
\centering
\small
    \begin{tabular}{l  l l l l l l}
    \toprule
        \textbf{order}  & \textbf{BERTScore} & \textbf{BLEU} & \textbf{METEOR} & \textbf{ROUGE-1} & \textbf{ROURGE-2} & \textbf{ROUGE-L} \\
        \cmidrule(lr){2-7}
        ascending (ours) & 0.881 & 0.057 & 0.229 & 0.284 & 0.096 & 0.228 \\ 
        descending & 0.861 & 0.003 & 0.082 & 0.095 & 0.012 & 0.085 \\ 
    \bottomrule
    \end{tabular}
    \caption{Results of different loop orders tested. 
    Let $t$ denote the current training stage, then the ascending order for the training stage $t$ is Task\textsubscript{$t$}, Task\textsubscript{$1$}, Task\textsubscript{$2$}, ..., Task\textsubscript{$t-1$}.
    The descending order follows for the same training stage $t$ the form Task\textsubscript{$t$}, Task\textsubscript{$t-1$}, Task\textsubscript{$t-2$}, ..., Task\textsubscript{$1$}.
    }
    \label{tab:cl_asc_VS_cl_desc}
\end{table*}

\begin{table*}[ht]
\centering
\small
    \begin{tabular}{l  cccccc  c}
    \toprule
        \textbf{model}  & \textbf{BERTScore} & \textbf{BLEU} & \textbf{METEOR} & \textbf{ROUGE-1} & \textbf{ROUGE-2} & \textbf{ROUGE-L} & \textbf{Time} \\ 
        \cmidrule(lr){2-7} \cmidrule(lr){8-8}
        BART & \textbf{0.881} & \textbf{0.061} & \textbf{0.231} & \textbf{0.286} & \textbf{0.100} & \textbf{0.233} & 0.75h \\ 
        T5 & \textbf{0.881} & 0.052 & 0.218 & 0.282 & 0.090 & 0.229 & 1.15h \\ 
        PEGASUS & 0.876 & 0.058 & 0.215 & 0.264 & 0.094 & 0.216 & 1h \\ 
        \bottomrule
    \end{tabular}
    \caption{
    Results of different models used. 
    The models were finetuned on Reddit TIFU without pre-finetuning and with full precision.
    Values in \textbf{bold} represent the highest results for a training scheme.}
    \label{tab:model_comparison_reddit}
\end{table*}
\begin{table*}[ht]
\centering
\small
    \begin{tabular}{l  cccccc  c}
    \toprule
        \textbf{model}  & \textbf{BERTScore} & \textbf{BLEU} & \textbf{METEOR} & \textbf{ROUGE-1} & \textbf{ROUGE-2} & \textbf{ROUGE-L} & \textbf{Time} \\ 
        \cmidrule(lr){2-7} \cmidrule(lr){8-8}
        BART & \textbf{0.864} & \textbf{0.129} & \textbf{0.306} & \textbf{0.444} & \textbf{0.168} & 0.267 & 13.5h \\ 
        T5 & \textbf{0.864} & 0.120 & 0.291 & 0.416 & 0.153 & \textbf{0.272} & 27.5h \\ 
        PEGASUS & 0.858 & 0.122 & 0.291 & 0.414 & 0.148 & 0.253 & 18.5h \\ 
        \bottomrule
    \end{tabular}
    \caption{
    Results of different models used. 
    The models were finetuned on arXiv without pre-finetuning and with full precision.
    Values in \textbf{bold} represent the highest results for a training scheme.}
    \label{tab:model_comparison_arxiv}
\end{table*}

\section{Extended Results}
\subsection{Extended Results on Reddit TIFU} \label{ap:complete_results_reddit}
\Crefrange{tab:RQ1_cMTL_red}{tab:RQ4_sim_wo_red} show the detailed evaluation for each research question and all tested combinations of task families evaluated on the Reddit TIFU datasets. The tables are divided according to their training scheme, i.e., each table shows one of the three training schemes (sim, seq, cMTL).

\begin{table*}[htbp]
\small
  \centering
    \begin{tabular}{l ccc ccc}
    \toprule
   \textbf{ Task Families} & \textbf{BERTScore} & \textbf{BLEU} & \textbf{METEOR} & \textbf{ROUGE-1} & \textbf{ROUGE-2} & \textbf{ROUGE-L} \\
\cmidrule{2-7}    
    CLS   & 0.881 & 0.057 & 0.226 & 0.282 & 0.097 & 0.229 \\
    CMNS  & 0.881 & 0.055 & 0.226 & 0.282 & 0.095 & 0.228 \\
    NLI   & 0.869 & 0.000 & 0.030 & 0.088 & 0.006 & 0.083 \\
    RC    &  \underline{\textbf{0.882}} & 0.057 & 0.230 & 0.285 & \textbf{0.098} & 0.230 \\
    RC\textsuperscript{+}   & 0.881 & 0.056 & 0.224 & 0.281 & 0.096 & 0.229 \\
    SUM   & 0.881 & \textbf{0.061} & \textbf{0.231} & \textbf{0.287} & \textbf{0.098} & \textbf{0.231} \\
    \midrule
    BART  & 0.858 & 0.003 & 0.087 & 0.105 & 0.011 & 0.090 \\
    \bottomrule
    \end{tabular}%
  \caption{RQ1 results (single task family) for Reddit TIFU and the sequential strategy. Values in \textbf{bold} represent the highest results for a training scheme. 
  \underline{Underlined} values are the highest results for that dataset independent of training.}
  \label{tab:RQ1_seq_red}%
\end{table*}%

\begin{table*}[htbp]
\small
  \centering
    \begin{tabular}{l ccc ccc}
    \toprule
   \textbf{ Task Families} & \textbf{BERTScore} & \textbf{BLEU} & \textbf{METEOR} & \textbf{ROUGE-1} & \textbf{ROUGE-2} & \textbf{ROUGE-L} \\
\cmidrule{2-7}    
    CLS   & 0.881 & 0.061 & 0.233 & 0.286 & 0.099 & 0.232 \\
    CMNS  & 0.863 & 0.003 & 0.078 & 0.091 & 0.013 & 0.081 \\
    NLI   & 0.863 & 0.003 & 0.082 & 0.095 & 0.012 & 0.085 \\
    RC    & 0.881 & 0.061 & \textbf{0.235} & \underline{\textbf{0.290}} & 0.100 & 0.232 \\
    RC\textsuperscript{+}   & 0.863 & 0.003 & 0.082 & 0.095 & 0.012 & 0.085 \\
    SUM   &  \underline{\textbf{0.882}} & \textbf{0.062} & \textbf{0.235} & 0.288 & \textbf{0.102} & \textbf{0.234} \\
    \midrule
    BART  & 0.858 & 0.003 & 0.087 & 0.105 & 0.011 & 0.090 \\
    \bottomrule
    \end{tabular}%
  \caption{RQ1 results (single task family) for Reddit TIFU and the simultaneous strategy.
  Values in \textbf{bold} represent the highest results for a training scheme. 
  \underline{Underlined} values are the highest results for that dataset independent of training.}
  \label{tab:RQ1_sim_red}%
\end{table*}%

\begin{table*}[htbp]
\small
  \centering
    \begin{tabular}{l ccc ccc}
    \toprule
    \textbf{Task Families} & \textbf{BERTScore} & \textbf{BLEU}  & \textbf{METEOR} & \textbf{ROUGE-1} & \textbf{ROUGE-2} & \textbf{ROUGE-L} \\
\cmidrule{2-7}    
    CLS   & 0.853 & 0.002 & 0.060 & 0.095 & 0.012 & 0.085 \\
    CMNS  & 0.863 & 0.003 & 0.078 & 0.091 & 0.013 & 0.081 \\
    NLI   & 0.863 & 0.003 & 0.082 & 0.095 & 0.012 & 0.085 \\
    RC    & \textbf{0.881} & \textbf{0.059} & 0.230 & \textbf{0.287} & \textbf{0.098} & 0.231 \\
    RC\textsuperscript{+}   & 0.863 & 0.003 & 0.078 & 0.091 & 0.013 & 0.080 \\
    SUM   & \textbf{0.881} & \textbf{0.059} & \textbf{0.231} & \textbf{0.287} & \textbf{0.098} & \textbf{0.232} \\
    \midrule
    BART  & 0.858 & 0.003 & 0.087 & 0.105 & 0.011 & 0.090 \\
    \bottomrule
    \end{tabular}%
  \caption{RQ1 results (single task family) for Reddit TIFU and the continual multi-task learning strategy.
  Values in \textbf{bold} represent the highest results for a training scheme. 
  \underline{Underlined} values are the highest results for that dataset independent of training.}
  \label{tab:RQ1_cMTL_red}%
\end{table*}%

\begin{table*}[htbp]
\small
  \centering
    \begin{tabular}{l ccc ccc}
    \toprule
   \textbf{ Task Families} & \textbf{BERTScore} & \textbf{BLEU} & \textbf{METEOR} & \textbf{ROUGE-1} & \textbf{ROUGE-2} & \textbf{ROUGE-L} \\
\cmidrule{2-7}    
    ALL   & \textbf{0.880} & \textbf{0.053} & \textbf{0.222} & \textbf{0.278} & \textbf{0.092} & \textbf{0.225} \\
    \midrule
    BART  & 0.858 & 0.003 & 0.087 & 0.105 & 0.011 & 0.090 \\
    \bottomrule
    \end{tabular}%
  \caption{RQ1 results (all task families) for Reddit TIFU and the sequential strategy.
  Values in \textbf{bold} represent the highest results for a training scheme. 
  \underline{Underlined} values are the highest results for that dataset independent of training.}
  \label{tab:RQ1_seq_ALL_red}%
\end{table*}%

\begin{table*}[htbp]
\small
  \centering
    \begin{tabular}{l ccc ccc}
    \toprule
   \textbf{ Task Families} & \textbf{BERTScore} & \textbf{BLEU} & \textbf{METEOR} & \textbf{ROUGE-1} & \textbf{ROUGE-2} & \textbf{ROUGE-L} \\
\cmidrule{2-7}    
    ALL   & \textbf{0.881} & \textbf{0.057} & \textbf{0.228} & \textbf{0.283} & \textbf{0.095} & \textbf{0.228} \\
    \midrule
    BART  & 0.858 & 0.003 & 0.087 & 0.105 & 0.011 & 0.090 \\
    \bottomrule
    \end{tabular}%
  \caption{RQ1 results (all task families) for Reddit TIFU and the simultaneous strategy.
  Values in \textbf{bold} represent the highest results for a training scheme. 
  \underline{Underlined} values are the highest results for that dataset independent of training.}
  \label{tab:RQ1_sim_ALL_red}%
\end{table*}%

\begin{table*}[htbp]
\small
  \centering
    \begin{tabular}{l ccc ccc}
    \toprule
   \textbf{ Task Families} & \textbf{BERTScore} & \textbf{BLEU} & \textbf{METEOR} & \textbf{ROUGE-1} & \textbf{ROUGE-2} & \textbf{ROUGE-L} \\
\cmidrule{2-7}    ALL   & 0.819 & 0.000 & 0.037 & 0.000 & 0.000 & 0.000 \\
    \midrule
    BART  & 0.858 & 0.003 & 0.087 & 0.105 & 0.011 & 0.090 \\
    \bottomrule
    \end{tabular}%
  \caption{RQ1 results (all task families) for Reddit TIFU and the continual multi-task learning strategy.
  Values in \textbf{bold} represent the highest results for a training scheme. 
  \underline{Underlined} values are the highest results for that dataset independent of training.}
  \label{tab:RQ1_cMTL_ALL_red}%
\end{table*}%

\begin{table*}[htbp]
\small
  \centering
    \begin{tabular}{l ccc ccc}
    \toprule
   \textbf{ Task Families} & \textbf{BERTScore} & \textbf{BLEU} & \textbf{METEOR} & \textbf{ROUGE-1} & \textbf{ROUGE-2} & \textbf{ROUGE-L} \\
\cmidrule{2-7}    
    SUM + CLS & 0.881 & \textbf{0.061} & 0.230 & 0.284 & 0.098 & 0.230 \\
    SUM + CMNS & 0.881 & 0.060 & 0.232 & 0.287 & 0.098 & 0.231 \\
    SUM + NLI & 0.881 & 0.053 & 0.223 & 0.280 & 0.094 & 0.225 \\
    SUM + RC & \underline{\textbf{0.882}} & \textbf{0.061} & \textbf{0.233} & \textbf{0.288} & \underline{\textbf{0.100}} & \textbf{0.235} \\
    SUM + RC\textsuperscript{+} & 0.881 & 0.060 & 0.230 & 0.285 & 0.098 & 0.232 \\
    \midrule
    BART  & 0.858 & 0.003 & 0.087 & 0.105 & 0.011 & 0.090 \\
    \bottomrule
    \end{tabular}%
  \caption{RQ2 results (pairing of the summarization task family with another task family) for Reddit TIFU and the sequential strategy.
  Values in \textbf{bold} represent the highest results for a training scheme. 
  \underline{Underlined} values are the highest results for that dataset independent of training.}
  \label{tab:RQ2_seq_red}%
\end{table*}%

\begin{table*}[htbp]
\small
  \centering
    \begin{tabular}{l ccc ccc}
    \toprule
   \textbf{ Task Families} & \textbf{BERTScore} & \textbf{BLEU} & \textbf{METEOR} & \textbf{ROUGE-1} & \textbf{ROUGE-2} & \textbf{ROUGE-L} \\
\cmidrule{2-7} 
    SUM + CLS & \textbf{0.881} & 0.061 & \textbf{0.233} & \textbf{0.287} & 0.096 & \textbf{0.232} \\
    SUM + CMNS & \textbf{0.881} & 0.059 & 0.231 & 0.284 & 0.097 & 0.230 \\
    SUM + NLI & \textbf{0.881} & \textbf{0.062} & \textbf{0.233} & \textbf{0.287} & \textbf{0.098} & 0.231 \\
    SUM + RC & \textbf{0.881} & 0.059 & 0.229 & 0.286 & 0.097 & 0.231 \\
    SUM + RC\textsuperscript{+} & \textbf{0.881} & 0.057 & 0.225 & 0.283 & 0.096 & 0.229 \\
    \midrule
    BART  & 0.858 & 0.003 & 0.087 & 0.105 & 0.011 & 0.090 \\
    \bottomrule
    \end{tabular}%
  \caption{RQ2 results (pairing of the summarization task family with another task family) for Reddit TIFU and the simultaneous strategy.
  Values in \textbf{bold} represent the highest results for a training scheme. 
  \underline{Underlined} values are the highest results for that dataset independent of training.}
  \label{tab:RQ2_sim_red}%
\end{table*}%

\begin{table*}[htbp]
\small
  \centering
    \begin{tabular}{l ccc ccc}
    \toprule
   \textbf{ Task Families} & \textbf{BERTScore} & \textbf{BLEU} & \textbf{METEOR} & \textbf{ROUGE-1} & \textbf{ROUGE-2} & \textbf{ROUGE-L} \\
\cmidrule{2-7}
    SUM + CLS & 0.864 & 0.003 & 0.077 & 0.093 & 0.013 & 0.081 \\
    SUM + CMNS & \textbf{0.881} & \textbf{0.062} & 0.234 & 0.289 & \underline{\textbf{0.100}} & \underline{\textbf{0.236}} \\
    SUM + NLI & \textbf{0.881} & 0.053 & 0.223 & 0.280 & 0.095 & 0.225 \\
    SUM + RC & \textbf{0.881} & \textbf{0.062} & \textbf{0.234} & \underline{\textbf{0.290}} & \underline{\textbf{0.100}} & 0.233 \\
    SUM + RC\textsuperscript{+} & \textbf{0.881} & 0.061 & 0.234 & 0.288 & \underline{\textbf{0.100}} & 0.233 \\
    \midrule
    BART  & 0.858 & 0.003 & 0.087 & 0.105 & 0.011 & 0.090 \\
    \bottomrule
    \end{tabular}%
  \caption{RQ2 results (pairing of the summarization task family with another task family) for Reddit TIFU and the continual multi-task learning strategy.
  Values in \textbf{bold} represent the highest results for a training scheme. 
  \underline{Underlined} values are the highest results for that dataset independent of training.}
  \label{tab:RQ2_cMTL_red}%
\end{table*}%

\begin{table*}[htbp]
\small
  \centering
    \begin{tabular}{l ccc ccc}
    \toprule
    \textbf{ Task Families} & \textbf{BERTScore} & \textbf{BLEU} & \textbf{METEOR} & \textbf{ROUGE-1} & \textbf{ROUGE-2} & \textbf{ROUGE-L} \\
\cmidrule{2-7}    
        CLS + CMNS & 0.863 & 0.003 & 0.078 & 0.091 & 0.013 & 0.081 \\ 
        CLS + NLI & 0.864 & 0.003 & 0.077 & 0.093 & 0.013 & 0.081 \\ 
        CLS + RC & \textbf{0.881} & 0.059 & 0.231 & 0.288 & 0.097 & 0.232 \\ 
        CLS + RC\textsuperscript{+} & \textbf{0.881} & 0.059 & 0.229 & 0.286 & 0.097 & 0.231 \\
        CMNS + NLI & \textbf{0.881} & 0.060 & 0.231 & 0.286 & 0.099 & 0.231 \\ 
        CMNS + RC & \textbf{0.881} & 0.059 & 0.227 & 0.282 & 0.096 & 0.228 \\ 
        CMNS + RC\textsuperscript{+} & \textbf{0.881} & \textbf{0.061} & 0.232 & 0.287 & 0.097 & 0.231 \\ 
        NLI + RC\textsuperscript{+} & \textbf{0.881} & \textbf{0.061} & \textbf{0.233} & \textbf{0.289} & \underline{\textbf{0.100}} & \textbf{0.234} \\ 
        NLI + RC & \textbf{0.881} & 0.058 & 0.231 & 0.286 & 0.097 & 0.231 \\ 
        RC + RC\textsuperscript{+} & \textbf{0.881} & 0.058 & 0.228 & 0.284 & 0.096 & 0.230 \\ 
        \midrule
    BART  & 0.858 & 0.003 & 0.087 & 0.105 & 0.011 & 0.090 \\
    \bottomrule
    \end{tabular}%
  \caption{RQ3 results (pairing of two task families excluding the text summarization family) for Reddit TIFU and the sequential strategy.
  Values in \textbf{bold} represent the highest results for a training scheme. 
  \underline{Underlined} values are the highest results for that dataset, independent of training.}
\label{tab:RQ3_seq_red}%
\end{table*}%

\begin{table*}[htbp]
\small
  \centering
    \begin{tabular}{l ccc ccc}
    \toprule
   \textbf{ Task Families} & \textbf{BERTScore} & \textbf{BLEU} & \textbf{METEOR} & \textbf{ROUGE-1} & \textbf{ROUGE-2} & \textbf{ROUGE-L} \\
    \cmidrule{2-7}    
        CLS + CMNS & 0.863 & 0.003 & 0.078 & 0.091 & 0.013 & 0.081 \\ 
        CLS + NLI & 0.864 & 0.003 & 0.077 & 0.093 & 0.013 & 0.081 \\ 
        CLS + RC & \textbf{0.881} & 0.059 & 0.231 & 0.288 & 0.097 & 0.232 \\ 
        CLS + RC\textsuperscript{+} & \textbf{0.881} & 0.059 & 0.229 & 0.286 & 0.097 & 0.231 \\ 
        CMNS + NLI & \textbf{0.881} & 0.060 & 0.231 & 0.286 & 0.099 & 0.231 \\ 
        CMNS + RC & \textbf{0.881} & 0.059 & 0.227 & 0.282 & 0.096 & 0.228 \\ 
        CMNS + RC\textsuperscript{+} & \textbf{0.881} & \textbf{0.061} & 0.232 & 0.287 & 0.097 & 0.231 \\ 
        NLI + RC & \textbf{0.881} & 0.058 & 0.231 & 0.286 & 0.097 & 0.231 \\ 
        NLI + RC\textsuperscript{+} & \textbf{0.881} & \textbf{0.061} & \textbf{0.234} & \textbf{0.289} & \underline{\textbf{0.100}} & \textbf{0.234} \\ 
        RC + RC\textsuperscript{+} & \textbf{0.881} & 0.058 & 0.228 & 0.284 & 0.096 & 0.223 \\ 
    \midrule
    BART  & 0.858 & 0.003 & 0.087 & 0.105 & 0.011 & 0.090 \\
    \bottomrule
    \end{tabular}%
  \caption{RQ3 results (pairing of two task families excluding the text summarization family) for Reddit TIFU and the simultaneous strategy.
  Values in \textbf{bold} represent the highest results for a training scheme. 
  \underline{Underlined} values are the highest results for that dataset independent of training.}
  \label{tab:RQ3_sim_red}%
\end{table*}%

\begin{table*}[htbp]
\small
  \centering
    \begin{tabular}{l ccc ccc}
    \toprule
   \textbf{ Task Families} & \textbf{BERTScore} & \textbf{BLEU} & \textbf{METEOR} & \textbf{ROUGE-1} & \textbf{ROUGE-2} & \textbf{ROUGE-L} \\
\cmidrule{2-7}
    CLS + CMNS & 0.853 & 0.002 & 0.060 & 0.095 & 0.012 & 0.085 \\
    CLS + NLI & 0.869 & 0.000 & 0.046 & 0.056 & 0.007 & 0.055 \\
    CLS + RC & \textbf{0.881} & 0.060 & 0.230 & 0.286 & 0.099 & 0.232 \\
    CLS + RC\textsuperscript{+} & 0.863 & 0.003 & 0.082 & 0.095 & 0.012 & 0.085 \\
    CMNS + NLI & 0.865 & 0.002 & 0.081 & 0.099 & 0.012 & 0.089 \\
    CMNS + RC & 0.864 & 0.003 & 0.077 & 0.093 & 0.013 & 0.081 \\
    CMNS + RC\textsuperscript{+} & \textbf{0.881} & \textbf{0.062} & \textbf{0.232} & \textbf{0.287} & \textbf{0.099} & \textbf{0.233} \\
    NLI + RC & \textbf{0.881} & 0.060 & 0.231 & 0.287 & 0.098 & 0.232 \\
    NLI + RC\textsuperscript{+} & \textbf{0.881} & 0.057 & 0.227 & 0.283 & 0.096 & 0.229 \\
    RC + RC\textsuperscript{+} & \textbf{0.881} & 0.059 & 0.228 & 0.284 & 0.098 & 0.230 \\
    \midrule
    BART  & 0.858 & 0.003 & 0.087 & 0.105 & 0.011 & 0.090 \\
    \bottomrule
    \end{tabular}%
  \caption{RQ3 results (pairing of two task families excluding the text summarization family) for Reddit TIFU and the continual multi-task learning strategy.
  Values in \textbf{bold} represent the highest results for a training scheme. 
  \underline{Underlined} values are the highest results for that dataset independent of training.}
  \label{tab:RQ3_cMTL_red}%
\end{table*}%

\begin{table*}[htbp]
\small
  \centering
    \begin{tabular}{l ccc ccc}
    \toprule
    \textbf{ Task Families}  & \textbf{BERTScore}  & \textbf{BLEU}  & \textbf{METEOR}  & \textbf{ROUGE-1}  & \textbf{ROUGE-2}  & \textbf{ROUGE-L} \\
    \cmidrule{2-7}    
    SUM + CLS + CMNS &  0.881  &  0.060  &  0.228  &  0.286  &  0.098  &  0.232 \\
    SUM + CLS + NLI &  0.881  &  0.059  &  0.231  &  0.285  &  0.098  &  0.231 \\
    SUM + CLS + RC &   \underline{\textbf{0.882}}  &  0.060  &  \textbf{0.235}  &  0.288  &  0.099  &  \textbf{0.234} \\
    SUM + CLS + RC\textsuperscript{+} & 0.881 &  \textbf{0.062}  &  \textbf{0.235}  &  0.288  &  \underline{\textbf{0.100}}  &  0.232 \\
    SUM + CMNS + NLI &  0.881  &  0.059  &  0.230  &  0.284  &  0.096  &  0.229 \\
    SUM + CMNS + RC &   \underline{\textbf{0.882}}  &  0.061  &  0.234  &  0.288  &  0.099  &  0.232 \\
    SUM + CMNS + RC\textsuperscript{+} &  0.881  &  \textbf{0.062}  &  0.232  &  0.287  &  \underline{\textbf{0.100}}  &  0.233 \\
    SUM + NLI + RC &  0.881  &  0.060  &  0.229  &  0.283  &  0.096  &  0.230 \\
    SUM + NLI + RC\textsuperscript{+} &  0.881  &  0.061  &  0.234  &  \textbf{0.289}  &  0.099  &  \textbf{0.234} \\
    SUM + RC + RC\textsuperscript{+} &  \underline{\textbf{0.882}}  &  0.058  &  0.227  &  0.284  &  0.099  &  0.232 \\
    \midrule
    BART  & 0.858 & 0.003 & 0.087 & 0.105 & 0.011 & 0.090 \\
    \bottomrule
    \end{tabular}%
  \caption{RQ4 results (pairing of the summarization task family with two other task families) for Reddit TIFU and the sequential strategy.
  Values in \textbf{bold} represent the highest results for a training scheme. 
  \underline{Underlined} values are the highest results for that dataset independent of training.}
  \label{tab:RQ4_seq_w_red}%
\end{table*}%

\begin{table*}[htbp]
\small
  \centering
    \begin{tabular}{l ccc ccc}
    \toprule
    \textbf{ Task Families}  & \textbf{BERTScore}  & \textbf{BLEU}  & \textbf{METEOR}  & \textbf{ROUGE-1}  & \textbf{ROUGE-2}  & \textbf{ROUGE-L} \\
    \cmidrule{2-7}    
    SUM + RC\textsuperscript{+} + CLS &  \textbf{0.881} &  0.061 &  0.233 &  0.289 &  0.099 &  0.232 \\
    SUM + RC\textsuperscript{+} + CMNS &  \textbf{0.881} &  0.061 &  0.231 &  0.286 &  0.099 &  0.232 \\
    SUM + RC\textsuperscript{+} + NLI &  \textbf{0.881} &  0.058 &  0.229 &  0.285 &  0.098 &  0.231 \\
    SUM + RC\textsuperscript{+} + RC &  \textbf{0.881} &  0.059 &  0.234 &  0.287 &  0.097 &  0.232 \\
    SUM + CLS + CMNS &  \textbf{0.881} &  0.057 &  0.227 &  0.283 &  0.096 &  0.229 \\
    SUM + CLS + NLI &  \textbf{0.881} &  0.060 &  0.231 &  0.284 &  0.099 &  0.229 \\
    SUM + CLS + RC &  \textbf{0.881} &  0.058 &  0.228 &  0.286 &  0.098 &  0.230 \\
    SUM + CMNS + NLI &  \textbf{0.881} &  \underline{\textbf{0.064}} &  \underline{\textbf{0.236}} &  \textbf{0.289} & \underline{\textbf{0.100}} &  \textbf{0.233} \\
    SUM + CMNS + RC &  \textbf{0.881} &  0.061 &  0.232 &  0.288 &  0.099 &  \textbf{0.233} \\
    SUM + NLI + RC &  \textbf{0.881} &  0.061 &  0.231 &  0.287 &  0.098 &  \textbf{0.233} \\
    \midrule
    BART  & 0.858 & 0.003 & 0.087 & 0.105 & 0.011 & 0.090 \\
    \bottomrule
    \end{tabular}%
  \caption{RQ4 results (pairing of the summarization task family with two other task families) for Reddit TIFU and the simultaneous strategy.
  Values in \textbf{bold} represent the highest results for a training scheme. 
  \underline{Underlined} values are the highest results for that dataset independent of training.}
  \label{tab:RQ4_sim_w_red}%
\end{table*}%

\begin{table*}[htbp]
\small
  \centering
    \begin{tabular}{l ccc ccc}
    \toprule
   \textbf{ Task Families} & \textbf{BERTScore} & \textbf{BLEU} & \textbf{METEOR} & \textbf{ROUGE-1} & \textbf{ROUGE-2} & \textbf{ROUGE-L} \\
    \cmidrule(lr){2-7}
    SUM + CLS + CMNS & 0.864 & 0.003 & 0.077 & 0.093 & 0.013 & 0.081 \\
    SUM + CLS + NLI & 0.863 & 0.003 & 0.082 & 0.095 & 0.012 & 0.085 \\
    SUM + CLS + RC & \textbf{0.881} & 0.058 & 0.229 & \textbf{0.285} & 0.098 & 0.231 \\
    SUM + CLS + RC\textsuperscript{+} & 0.863 & 0.003 & 0.082 & 0.095 & 0.012 & 0.085 \\
    SUM + CMNS + NLI & \textbf{0.881} & \textbf{0.059} & 0.229 & \textbf{0.285} & 0.098 & 0.230 \\
    SUM + CMNS + RC & \textbf{0.881} & \textbf{0.059} & \textbf{0.230} & \textbf{0.285} & \textbf{0.099} & \textbf{0.232} \\
    SUM + CMNS + RC\textsuperscript{+} & \textbf{0.881} & \textbf{0.059} & 0.228 & 0.284 & 0.096 & 0.229 \\
    SUM + NLI + RC & \textbf{0.881} & \textbf{0.059} & 0.228 & 0.284 & 0.096 & 0.230 \\
    SUM + NLI + RC\textsuperscript{+} & \textbf{0.881} & 0.058 & 0.229 & 0.284 & 0.096 & 0.230 \\
    SUM + RC + RC\textsuperscript{+} & \textbf{0.881} & \textbf{0.059} & 0.228 & \textbf{0.285} & 0.097 & 0.230 \\
    \midrule
    BART  & 0.858 & 0.003 & 0.087 & 0.105 & 0.011 & 0.090 \\
    \bottomrule
    \end{tabular}%
  \caption{RQ4 results (pairing of the summarization task family with two other task families) for Reddit TIFU and the contniual multi-task learning strategy.
  Values in \textbf{bold} represent the highest results for a training scheme. 
  \underline{Underlined} values are the highest results for that dataset independent of training.}
  \label{tab:RQ4_cMTL_w_red}%
\end{table*}%

\begin{table*}[htbp]
\small
  \centering
    \begin{tabular}{l ccc ccc}
    \toprule
    \textbf{ Task Families}  & \textbf{BERTScore}  & \textbf{BLEU}  & \textbf{METEOR}  & \textbf{ROUGE-1}  & \textbf{ROUGE-2}  & \textbf{ROUGE-L} \\
    \cmidrule{2-7}
    CLS + CMNS + NLI &  0.752 &   0.000 &   0.034 &   0.044 &   0.000 &   0.040 \\
    CLS + CMNS + RC &  \textbf{0.881} &   \textbf{0.062} &   \textbf{0.235} &   0.287 &   \textbf{0.099} &   0.231 \\
    CLS + CMNS + RC\textsuperscript{+} &  \textbf{0.881} &   \textbf{0.062} &   0.231 &   0.286 &   0.098 &   0.232 \\
    CLS + NLI + RC &  \textbf{0.881} &   0.059 &   0.233 &   \textbf{0.289} &   \textbf{0.099} &   \textbf{0.233} \\
    CLS + NLI + RC\textsuperscript{+} &  \textbf{0.881} &   0.059 &   0.232 &   0.286 &   0.097 &   0.231 \\
    CLS + RC + RC\textsuperscript{+} &  0.880 &   0.060 &   0.232 &   0.285 &   0.098 &   0.230 \\
    CMNS + NLI + RC &  0.880 &   0.059 &   0.229 &   0.284 &   0.095 &   0.230 \\
    CMNS + NLI + RC\textsuperscript{+} &  \textbf{0.881} &   0.059 &   0.231 &   0.284 &   0.096 &   0.230 \\
    CMNS + RC + RC\textsuperscript{+} &  \textbf{0.881} &   0.058 &   0.232 &   0.285 &   0.097 &   0.230 \\
    NLI + RC + RC\textsuperscript{+} &  \textbf{0.881} &   0.058 &   0.230 &   0.284 &   0.097 &   0.229 \\

    \midrule
    BART  & 0.858 & 0.003 & 0.087 & 0.105 & 0.011 & 0.090 \\
    \bottomrule
    \end{tabular}%
  \caption{RQ4 results  (pairing of three task families excluding the text summarization family) for Reddit TIFU and the sequential strategy.
  Values in \textbf{bold} represent the highest results for a training scheme. 
  \underline{Underlined} values are the highest results for that dataset independent of training.}
  \label{tab:RQ4_seq_wo_red}%
\end{table*}%

\begin{table*}[htbp]
\small
  \centering
    \begin{tabular}{l ccc ccc}
    \toprule
    \textbf{ Task Families}  & \textbf{BERTScore}  & \textbf{BLEU}  & \textbf{METEOR}  & \textbf{ROUGE-1}  & \textbf{ROUGE-2}  & \textbf{ROUGE-L}  \\
\cmidrule{2-7}    
    CLS + CMNS + NLI &  0.746  &  0.000  &  0.024  &  0.028  &  0.000  &  0.275  \\
    CLS + CMNS + RC &  \textbf{0.881}  &  0.060  &  0.232  &  0.287  &  \textbf{0.099}  &  \textbf{0.232}  \\
    CLS + CMNS + RC\textsuperscript{+} &  0.863  &  0.003  &  0.082  &  0.095  &  0.012  &  0.085  \\
    CLS + NLI + RC &  \textbf{0.881}  &  0.059  &  0.228  &  0.285  &  0.098  &  0.230  \\
    CLS + NLI + RC\textsuperscript{+} &  \textbf{0.881}  &  0.057  &  0.225  &  0.283  &  0.097  &  0.231  \\
    CLS + RC + RC\textsuperscript{+} &  \textbf{0.881}  &  0.058  &  0.227  &  0.282  &  0.097  &  0.229  \\
    CMNS + NLI + RC &  0.766  &  0.000  &  0.020  &  0.009  &  0.000  &  0.009  \\
    CMNS + NLI + RC\textsuperscript{+} &  \textbf{0.881}  &  0.058  &  0.230  &  0.283  &  0.097  &  0.229  \\
    CMNS + RC + RC\textsuperscript{+} &  \textbf{0.881}  &  \textbf{0.061}  &  \textbf{0.234}  &  \textbf{0.288}  &  0.097  &  0.231  \\
    NLI + RC + RC\textsuperscript{+} &  \textbf{0.881}  &  0.059  &  0.230  &  0.284  &  0.096  &  0.229  \\

    \midrule
    BART  & 0.858 & 0.003 & 0.087 & 0.105 & 0.011 & 0.090 \\
    \bottomrule
    \end{tabular}%
  \caption{RQ4 results  (pairing of three task families excluding the text summarization family) for Reddit TIFU and the simultaneous strategy.
  Values in \textbf{bold} represent the highest results for a training scheme. 
  \underline{Underlined} values are the highest results for that dataset independent of training.}
  \label{tab:RQ4_sim_wo_red}%
\end{table*}%

\begin{table*}[htbp]
\small
  \centering
    \begin{tabular}{l ccc ccc}
    \toprule
    \textbf{ Task Families}  & \textbf{BERTScore}  & \textbf{BLEU}  & \textbf{METEOR}  & \textbf{ROUGE-1}  & \textbf{ROUGE-2}  & \textbf{ROUGE-L} \\
    \cmidrule{2-7}
    
    CLS + CMNS + NLI &  0.751 &   0.000 &   0.017 &   0.000 &   0.000 &   0.000 \\
    CLS + CMNS + RC &  0.753 &   0.000 &   0.009 &   0.015 &   0.000 &   0.015 \\
    CLS + CMNS + RC\textsuperscript{+} &  0.861 &   0.002 &   0.064 &   0.057 &   0.012 &   0.054 \\
    CLS + NLI + RC &  0.864 &   0.003 &   0.077 &   0.093 &   0.013 &   0.081 \\
    CLS + NLI + RC\textsuperscript{+} &  0.863 &   0.003 &   0.082 &   0.095 &   0.012 &   0.085 \\
    CLS + RC + RC\textsuperscript{+}&  0.747 &   0.000 &   0.025 &   0.020 &   0.000 &   0.020 \\
    CMNS + NLI + RC &  0.867 &   0.004 &   0.105 &   0.125 &   0.012 &   0.101 \\
    CMNS + NLI + RC\textsuperscript{+} &  \textbf{0.881} &   0.058 &   0.228 &   0.285 &   0.096 &   0.230 \\
    CMNS + RC + RC\textsuperscript{+} &  \textbf{0.881} &   \textbf{0.060} &   0.229 &   0.284 &   \textbf{0.098} &   0.230 \\
    NLI + RC + RC\textsuperscript{+} &  \textbf{0.881} &   0.059 &   \textbf{0.231} &   \textbf{0.286} &   \textbf{0.098} &   \textbf{0.231} \\
    
    \midrule
    BART  & 0.858 & 0.003 & 0.087 & 0.105 & 0.011 & 0.090 \\
    \bottomrule
    \end{tabular}%
  \caption{RQ4 results  (pairing of three task families excluding the text summarization family) for Reddit TIFU and the continual multi-task learning strategy.
  Values in \textbf{bold} represent the highest results for a training scheme. 
  \underline{Underlined} values are the highest results for that dataset independent of training.}
  \label{tab:RQ4_cMTL_wo_red}%
\end{table*}%

\subsection{Extended Results on arXiv} \label{ap:complete_results_arxiv}
\Crefrange{tab:RQ1_cMTL_arx}{tab:RQ3_sim_arx} show the detailed evaluation for each research question and all tested combinations of task families evaluated on the arXiv datasets. The tables are divided according to their training scheme, i.e., each table shows one of the three training schemes (sim, seq, cMTL).
\begin{table*}[htbp]
\small
  \centering
    \begin{tabular}{l ccc ccc}
    \toprule
    \textbf{ Task Families}  & \textbf{BERTScore}  & \textbf{BLEU}  & \textbf{METEOR}  & \textbf{ROUGE-1}  & \textbf{ROUGE-2}  & \textbf{ROUGE-L}  \\
\cmidrule{2-7}    
    CLS   & 0.820 & 0.018 & 0.154 & 0.248 & 0.048 & 0.163 \\
    CMNS  & \textbf{0.860} & 0.119 & 0.286 & \underline{\textbf{0.432}} & \textbf{0.167} & \underline{\textbf{0.249}} \\
    NLI   & 0.817 & 0.020 & 0.168 & 0.266 & 0.048 & 0.169 \\
    RC    & 0.859 & 0.117 & 0.282 & 0.427 & 0.165 & 0.247 \\
    RC\textsuperscript{+}   & 0.859 & 0.117 & 0.282 & 0.426 & 0.164 & 0.246 \\
    SUM   & 0.859 & \textbf{0.121} & \textbf{0.288} & 0.431 & \textbf{0.167} & \underline{\textbf{0.249}} \\
    \midrule
    BART  & 0.859 & 0.116 & 0.281 & 0.425 & 0.163 & 0.246 \\
    \bottomrule
    \end{tabular}%
  \caption{RQ1 results (single task family) for arXiv and the sequential strategy.
  Values in \textbf{bold} represent the highest results for a training scheme. 
  \underline{Underlined} values are the highest results for that dataset independent of training.}
  \label{tab:RQ1_seq_arx}%
\end{table*}%

\begin{table*}[htbp]
\small
  \centering
    \begin{tabular}{l ccc ccc}
    \toprule
    \textbf{ Task Families}  & \textbf{BERTScore}  & \textbf{BLEU}  & \textbf{METEOR}  & \textbf{ROUGE-1}  & \textbf{ROUGE-2}  & \textbf{ROUGE-L}  \\
\cmidrule{2-7}    
    CLS   & \textbf{0.860} & \textbf{0.120} & 0.287 & 0.430 & \textbf{0.167} & \textbf{0.248} \\
    CMNS  & 0.806 & 0.011 & 0.197 & 0.215 & 0.038 & 0.137 \\
    NLI   & 0.812 & 0.006 & 0.111 & 0.187 & 0.016 & 0.123 \\
    RC    & 0.859 & 0.119 & 0.284 & 0.430 & 0.166 & \textbf{0.248} \\
    RC\textsuperscript{+}   & 0.859 & \textbf{0.120} & \textbf{0.289} & \textbf{0.431} & \textbf{0.167} & \textbf{0.248} \\
    SUM   & 0.859 & 0.117 & 0.282 & 0.429 & 0.166 & \textbf{0.248} \\
    \midrule
    BART  & 0.859 & 0.116 & 0.281 & 0.425 & 0.163 & 0.246 \\
    \bottomrule
    \end{tabular}%
  \caption{RQ1 results (single task family) for arXiv and the simultaneous strategy.
  Values in \textbf{bold} represent the highest results for a training scheme. 
  \underline{Underlined} values are the highest results for that dataset independent of training.}
  \label{tab:RQ1_sim_arx}%
\end{table*}%

\begin{table*}[htbp]
\small
  \centering
    \begin{tabular}{l ccc ccc}
    \toprule
    \textbf{ Task Families}  & \textbf{BERTScore}  & \textbf{BLEU}  & \textbf{METEOR}  & \textbf{ROUGE-1}  & \textbf{ROUGE-2}  & \textbf{ROUGE-L}  \\
\cmidrule(lr){2-2} \cmidrule(lr){3-3} \cmidrule(lr){4-4} \cmidrule(lr){5-7}    
    CLS   & 0.859 & \textbf{0.119} & \textbf{0.286} & 0.429 & 0.166 & 0.248 \\
    CMNS  & 0.819 & 0.017 & 0.163 & 0.295 & 0.051 & 0.171 \\
    NLI   & 0.815 & 0.018 & 0.182 & 0.272 & 0.044 & 0.170 \\
    RC    & 0.859 & 0.117 & 0.282 & 0.426 & 0.164 & 0.246 \\
    RC\textsuperscript{+}   & 0.817 & 0.020 & 0.203 & 0.249 & 0.051 & 0.159 \\
    SUM   & \textbf{0.860} & \textbf{0.119} & \textbf{0.286} & \textbf{0.431} & \textbf{0.167} & \underline{\textbf{0.249}} \\
    \midrule
    BART  & 0.859 & 0.116 & 0.281 & 0.425 & 0.163 & 0.246 \\
    \bottomrule
    \end{tabular}%
  \caption{RQ1 results (single task family) for arXiv and the continual multi-task learning strategy.
  Values in \textbf{bold} represent the highest results for a training scheme. 
  \underline{Underlined} values are the highest results for that dataset independent of training.}
  \label{tab:RQ1_cMTL_arx}%
\end{table*}%

\begin{table*}[htbp]
\small
  \centering
    \begin{tabular}{l ccc ccc}
    \toprule
    \textbf{ Task Families}  & \textbf{BERTScore}  & \textbf{BLEU}  & \textbf{METEOR}  & \textbf{ROUGE-1}  & \textbf{ROUGE-2}  & \textbf{ROUGE-L}  \\
\cmidrule{2-7}    
    ALL   & 0.859 & 0.116 & 0.281 & \textbf{0.427} & \textbf{0.165} & \textbf{0.248} \\
    \midrule
    BART  & 0.859 & 0.116 & 0.281 & 0.425 & 0.163 & 0.246 \\
    \bottomrule
    \end{tabular}%
  \caption{RQ1 results (all task families) for arXiv and the sequential strategy.
  Values in \textbf{bold} represent the highest results for a training scheme. 
  \underline{Underlined} values are the highest results for that dataset independent of training.}
  \label{tab:RQ1_seq_ALL_arx}%
\end{table*}%

\begin{table*}[htbp]
\small
  \centering
    \begin{tabular}{l ccc ccc}
    \toprule
    \textbf{ Task Families}  & \textbf{BERTScore}  & \textbf{BLEU}  & \textbf{METEOR}  & \textbf{ROUGE-1}  & \textbf{ROUGE-2}  & \textbf{ROUGE-L}  \\
\cmidrule{2-7}    
    ALL   & 0.859 & 0.115 & 0.279 & 0.425 & \textbf{0.164} & 0.246 \\
    \midrule
    BART  & 0.859 & 0.116 & 0.281 & 0.425 & 0.163 & 0.246 \\
    \bottomrule
    \end{tabular}%
  \caption{RQ1 results (all task families) for arXiv and the simultaneous strategy.
  Values in \textbf{bold} represent the highest results for a training scheme. 
  \underline{Underlined} values are the highest results for that dataset independent of training.}
  \label{tab:RQ1_sim_ALL_arx}%
\end{table*}%

\begin{table*}[htbp]
\small
  \centering
    \begin{tabular}{l ccc ccc}
    \toprule
    \textbf{ Task Families}  & \textbf{BERTScore}  & \textbf{BLEU}  & \textbf{METEOR}  & \textbf{ROUGE-1}  & \textbf{ROUGE-2}  & \textbf{ROUGE-L}  \\
\cmidrule{2-7}  
    ALL   & 0.729 & 0.000 & 0.008 & 0.009 & 0.000 & 0.009 \\
    \midrule
    BART  & 0.859 & 0.116 & 0.281 & 0.425 & 0.163 & 0.246 \\
    \bottomrule
    \end{tabular}%
  \caption{RQ1 results (all task families) for arXiv and the continual multi-task learning strategy.
  Values in \textbf{bold} represent the highest results for a training scheme. 
  \underline{Underlined} values are the highest results for that dataset independent of training.}
  \label{tab:RQ1_cMTL_ALL_arx}%
\end{table*}%

\begin{table*}[htbp]
\small
  \centering
    \begin{tabular}{l ccc ccc}
    \toprule
    \textbf{ Task Families}  & \textbf{BERTScore}  & \textbf{BLEU}  & \textbf{METEOR}  & \textbf{ROUGE-1}  & \textbf{ROUGE-2}  & \textbf{ROUGE-L}  \\
\cmidrule{2-7}    
    SUM + CLS & \textbf{0.860} & \textbf{0.119} & \textbf{0.285} & \textbf{0.430} & \textbf{0.167} & \underline{\textbf{0.249}} \\
    SUM + CMNS & 0.811 & 0.016 & 0.153 & 0.254 & 0.046 & 0.164 \\
    SUM + NLI & 0.859 & 0.117 & 0.282 & 0.427 & 0.165 & 0.247 \\
    SUM + RC & 0.859 & \textbf{0.119} & \textbf{0.285} & \textbf{0.430} & 0.166 & 0.248 \\
    SUM + RC\textsuperscript{+} & 0.859 & 0.118 & 0.284 & 0.428 & 0.166 & 0.247 \\
    \midrule
    BART  & 0.859 & 0.116 & 0.281 & 0.425 & 0.163 & 0.246 \\
    \bottomrule
    \end{tabular}%
  \caption{RQ2 results (pairing of the summarization task family with another task family) for arXiv and the sequential strategy.
  Values in \textbf{bold} represent the highest results for a training scheme. 
  \underline{Underlined} values are the highest results for that dataset independent of training.}
  \label{tab:RQ2_seq_arx}%
\end{table*}%

\begin{table*}[htbp]
\small
  \centering
    \begin{tabular}{l ccc ccc}
    \toprule
    \textbf{ Task Families}  & \textbf{BERTScore}  & \textbf{BLEU}  & \textbf{METEOR}  & \textbf{ROUGE-1}  & \textbf{ROUGE-2}  & \textbf{ROUGE-L}  \\
\cmidrule{2-7}    
    SUM + CLS & \textbf{0.860} & 0.119 & 0.285 & 0.429 & 0.166 & 0.247 \\
    SUM + CMNS & \textbf{0.860} & 0.119 & 0.286 & \underline{\textbf{0.432}} & \textbf{0.167} & \underline{\textbf{0.249}} \\
    SUM + NLI & 0.859 & \textbf{0.120} & \textbf{0.287} & 0.431 & \textbf{0.167} & \underline{\textbf{0.249}} \\
    SUM + RC & 0.859 & 0.115 & 0.280 & 0.427 & 0.164 & 0.247 \\
    SUM + RC\textsuperscript{+} & 0.859 & 0.116 & 0.281 & 0.427 & 0.164 & 0.247 \\
    \midrule
    BART  & 0.859 & 0.116 & 0.281 & 0.425 & 0.163 & 0.246 \\
    \bottomrule
    \end{tabular}%
    \caption{RQ2 results (pairing of the summarization task family with another task family) for arXiv and the simultaneous strategy.
  Values in \textbf{bold} represent the highest results for a training scheme. 
  \underline{Underlined} values are the highest results for that dataset independent of training.}
  \label{tab:RQ2_sim_arx}%
\end{table*}%

\begin{table*}[htbp]
\small
  \centering
    \begin{tabular}{l ccc ccc}
    \toprule
    \textbf{ Task Families}  & \textbf{BERTScore}  & \textbf{BLEU}  & \textbf{METEOR}  & \textbf{ROUGE-1}  & \textbf{ROUGE-2}  & \textbf{ROUGE-L}  \\
    \cmidrule{2-7}    
    SUM + CLS & 0.859 & 0.117 & 0.283 & 0.429 & 0.165 & 0.248 \\
    SUM + CMNS & \textbf{0.860} & \textbf{0.120} & \textbf{0.288} & \underline{\textbf{0.432}} & \textbf{0.167} & \underline{\textbf{0.249}} \\
    SUM + NLI & 0.859 & 0.117 & 0.282 & 0.427 & 0.165 & 0.247 \\
    SUM + RC & 0.859 & 0.118 & 0.283 & 0.428 & 0.166 & 0.247 \\
    SUM + RC\textsuperscript{+} & 0.859 & 0.118 & 0.284 & 0.428 & 0.166 & 0.247 \\
    \midrule
    BART  & 0.859 & 0.116 & 0.281 & 0.425 & 0.163 & 0.246 \\
    \bottomrule
    \end{tabular}%
  \caption{RQ2 results (pairing of the summarization task family with another task family) for arXiv and the continual multi-task learning strategy.
  Values in \textbf{bold} represent the highest results for a training scheme. 
  \underline{Underlined} values are the highest results for that dataset independent of training.}
  \label{tab:RQ2_cMTL_arx}%
\end{table*}%

\begin{table*}[htbp]
\small
  \centering
    \begin{tabular}{l ccc ccc}
    \toprule
    \textbf{ Task Families}  & \textbf{BERTScore}  & \textbf{BLEU}  & \textbf{METEOR}  & \textbf{ROUGE-1}  & \textbf{ROUGE-2}  & \textbf{ROUGE-L}  \\
\cmidrule{2-7}    
    CLS + CMNS & \underline{\textbf{0.863}} & 0.003 & 0.078 & 0.091 & 0.013 & 0.081 \\
    CLS + NLI & 0.731 & 0.000 & 0.050 & 0.086 & 0.000 & 0.050 \\
    CLS + RC & 0.859 & 0.116 & \textbf{0.287} & 0.427 & 0.165 & 0.247 \\
    CLS + RC\textsuperscript{+} & 0.859 & 0.118 & 0.284 & 0.430 & \textbf{0.167} & \textbf{0.248} \\
    CMNS + NLI & 0.821 & 0.010 & 0.137 & 0.261 & 0.045 & 0.176 \\
    CMNS + RC & 0.860 & 0.117 & 0.283 & 0.429 & 0.165 & \textbf{0.248} \\
    CMNS + RC\textsuperscript{+} & 0.859 & 0.115 & 0.279 & 0.426 & 0.164 & 0.247 \\
    NLI + RC & 0.859 & 0.119 & 0.285 & 0.429 & 0.166 & \textbf{0.248} \\
    NLI + RC\textsuperscript{+} & 0.859 & \textbf{0.119} & 0.286 & \textbf{0.431} & \textbf{0.167} & \textbf{0.248} \\
    RC + RC\textsuperscript{+} & 0.859 & 0.116 & \textbf{0.287} & 0.428 & 0.165 & \textbf{0.248} \\
    \midrule
    BART  & 0.859 & 0.116 & 0.281 & 0.425 & 0.163 & 0.246 \\
    \bottomrule
    \end{tabular}%
  \caption{RQ3 results (pairing of two task families excluding the text summarization family) for arXiv and the sequential strategy.
  Values in \textbf{bold} represent the highest results for a training scheme. 
  \underline{Underlined} values are the highest results for that dataset independent of training.}
  \label{tab:RQ3_seq_arx}%
\end{table*}%

\begin{table*}[htbp]
\small
  \centering
    \begin{tabular}{l ccc ccc}
    \toprule
    \textbf{ Task Families}  & \textbf{BERTScore}  & \textbf{BLEU}  & \textbf{METEOR}  & \textbf{ROUGE-1}  & \textbf{ROUGE-2}  & \textbf{ROUGE-L}  \\
\cmidrule{2-7}
    CLS + CMNS & 0.704 & 0.000 & 0.050 & 0.076 & 0.000 & 0.046 \\
    CLS + NLI & 0.743 & 0.000 & 0.003 & 0.006 & 0.000 & 0.006 \\
    CLS + RC & 0.859 & 0.118 & 0.283 & 0.428 & 0.165 & 0.247 \\
    CLS + RC\textsuperscript{+} & 0.859 & 0.120 & 0.288 & \underline{\textbf{0.432}} & 0.167 & 0.248 \\
    CMNS + NLI & 0.805 & 0.012 & 0.212 & 0.215 & 0.041 & 0.134 \\
    CMNS + RC & 0.859 & 0.118 & 0.284 & 0.428 & 0.165 & 0.248 \\
    CMNS + RC\textsuperscript{+} & 0.859 & 0.115 & 0.280 & 0.426 & 0.165 & 0.247 \\
    NLI + RC & 0.859 & 0.119 & 0.285 & 0.430 & 0.166 & 0.248 \\
    NLI + RC\textsuperscript{+} & 0.859 & \underline{\textbf{0.121}} & \underline{\textbf{0.290}} & \underline{\textbf{0.432}} & \underline{\textbf{0.168}} & \underline{\textbf{0.249}} \\
    RC + RC\textsuperscript{+} & 0.859 & 0.116 & 0.281 & 0.426 & 0.164 & 0.247 \\
    \midrule
    BART  & 0.859 & 0.116 & 0.281 & 0.425 & 0.163 & 0.246 \\
    \bottomrule
    \end{tabular}%
  \caption{RQ3 results (pairing of two task families excluding the text summarization family) for arXiv and the simultaneous strategy.
  Values in \textbf{bold} represent the highest results for a training scheme. 
  \underline{Underlined} values are the highest results for that dataset independent of training.}
  \label{tab:RQ3_sim_arx}%
\end{table*}%

\begin{table*}[htbp]
\small
  \centering
    \begin{tabular}{l ccc ccc}
    \toprule
    \textbf{ Task Families}  & \textbf{BERTScore}  & \textbf{BLEU}  & \textbf{METEOR}  & \textbf{ROUGE-1}  & \textbf{ROUGE-2}  & \textbf{ROUGE-L}  \\
\cmidrule{2-7}
    CLS + CMNS & 0.813 & 0.018 & 0.162 & 0.259 & 0.052 & 0.176 \\
    CLS + NLI & 0.859 & 0.113 & 0.276 & 0.422 & 0.161 & 0.245 \\
    CLS + RC & 0.810 & 0.018 & 0.181 & 0.269 & 0.048 & 0.168 \\
    CLS + RC\textsuperscript{+} & 0.860 & \textbf{0.120} & \textbf{0.287} & \textbf{0.432} & \textbf{0.167} & \underline{\textbf{0.249}} \\
    CMNS + NLI & 0.806 & 0.009 & 0.118 & 0.181 & 0.016 & 0.117 \\
    CMNS + RC & 0.812 & 0.019 & 0.179 & 0.282 & 0.041 & 0.157 \\
    CMNS + RC\textsuperscript{+} & \underline{\textbf{0.863}} & 0.003 & 0.082 & 0.095 & 0.117 & 0.085 \\
    NLI + RC & 0.860 & 0.118 & 0.284 & 0.429 & 0.166 & 0.247 \\
    NLI + RC\textsuperscript{+} & 0.859 & 0.117 & 0.282 & 0.426 & 0.164 & 0.246 \\
    RC + RC\textsuperscript{+} & 0.859 & 0.118 & 0.285 & 0.429 & 0.165 & 0.248 \\
    \midrule
    BART  & 0.859 & 0.116 & 0.281 & 0.425 & 0.163 & 0.246 \\
    \bottomrule
    \end{tabular}%
  \caption{RQ3 results (pairing of two task families excluding the text summarization family) for arXiv and the continual multi-task learning strategy.
  Values in \textbf{bold} represent the highest results for a training scheme. 
  \underline{Underlined} values are the highest results for that dataset independent of training.}
  \label{tab:RQ3_cMTL_arx}%
\end{table*}%

\section{Hyperparameters} \label{ap:hyperparameters}
\Cref{tab:hyperparameters} shows the hyperparameters used throughout the pre-finetuning and finetuning experiments.

\begin{table*}[htbp]
    \centering
    \begin{tabular}{ll}
        \hline
        \textbf{Hyper parameter} & \textbf{Value }           \\ \hline
        Optimizer       & AdamW            \\
        Adam-betas      & (0.9, 0.999)     \\
        Adam-eps        & 1e-8             \\
        LR              & 5e-05            \\
        LR Scheduler    & linear decay     \\
        Dropout         & 0.1              \\
        Weight Decay    & 0                \\
        Warmup Updates  & 0                \\\hline
    \end{tabular}
    \caption{Hyperparameters used throughout all pre-finetuning and finetuning experiments.}
    \label{tab:hyperparameters}
\end{table*}

\end{document}